\documentclass{article}%
\usepackage{amsmath}
\usepackage{amsfonts}
\usepackage{amssymb}
\usepackage{graphicx}
\usepackage{algorithmic}
\usepackage{algorithm}%
\providecommand{\U}[1]{\protect \rule{.1in}{.1in}}

\begin{document}

\title{Dual feature-based and example-based explanation methods}
\author{Andrei V. Konstantinov, Boris V. Kozlov, Stanislav R. Kirpichenko and Lev V.
Utkin \\Peter the Great St.Petersburg Polytechnic University \\St.Petersburg, Russia \\e-mail: andrue.konst@gmail.com, kozlov.99260422@gmail.com, \\kirpichenko.sr@edu.spbstu.ru, lev.utkin@gmail.com}
\date{}
\maketitle

\begin{abstract}
A new approach to the local and global explanation is proposed. It is based on
selecting a convex hull constructed for the finite number of points around an
explained instance. The convex hull allows us to consider a dual
representation of instances in the form of convex combinations of extreme
points of a produced polytope. Instead of perturbing new instances in the
Euclidean feature space, vectors of convex combination coefficients are
uniformly generated from the unit simplex, and they form a new dual dataset. A
dual linear surrogate model is trained on the dual dataset. The explanation
feature importance values are computed by means of simple matrix calculations.
The approach can be regarded as a modification of the well-known model LIME.
The dual representation inherently allows us to get the example-based
explanation. The neural additive model is also considered as a tool for
implementing the example-based explanation approach. Many numerical
experiments with real datasets are performed for studying the approach. The
code of proposed algorithms is available.


\end{abstract}

\section{Introduction}

The black-box nature of many machine learning models, including neural
networks, requires to explain their predictions in order to provide confidence
in them. This requirement affects many applications, especially those in
medicine, finance, safety maintenance. As a result, many successful methods
and algorithms have been developed to satisfy this requirement
\cite{Arya-etal-2019,Belle-Papantonis-2020,Guidotti-2019,Liang-etal-2021,Molnar-2019,Murdoch-etal-2019,Xie-Ras-etal-2020,Zablocki-etal-21,Zhang-Tino-etal-2020}%
.

There are many definitions and interpretations of the explanation. We
understand explanation as an answer to the question which features of an
instance or a set of instances significantly impact the black-box model
prediction or which features are most relevant to the prediction. Methods
answering this question can be referred to as \textit{feature importance}
methods or the \textit{feature-based explanation}. Another group of
explanation methods is called the \textit{example-based} explanation methods
\cite{Molnar-2019}. The corresponding methods are based on selecting
influential instances from a training set having the largest impact on
predictions to compare the training instance with the explainable one.

Feature importance explanation methods, in turn, can be divided into two
groups: local and global. Methods from the first group explain the black-box
model predictions locally around a test instance. Global methods explain a set
of instances or the entire dataset. The well-known local explanation method is
the Local Interpretable Model-Agnostic Explanation (LIME)
\cite{Ribeiro-etal-2016}. In accordance with this method, a surrogate model is
constructed and trained, which approximates the black-box model at a point.
The surrogate model in LIME is the linear regression whose coefficients can be
interpreted as the feature importance measures. In fact, LIME can be regarded
as a method of the linear approximation of a complex non-linear function
implemented by the black-box model at a point. LIME is based on using a simple
regression model. Agarval et al. \cite{Agarwal-etal-20} proposed to generalize
LIME using the generalized additive model (GAM) \cite{Hastie-Tibshirani-1990}
instead of the simple linear regression and its implementation by means of
neural networks of a special form. The GAM is a more general and flexible
model in comparison with the original linear model. The corresponding
surrogate model using the GAM is called the neural additive model (NAM).%

\begin{figure}
[ptb]
\begin{center}
\includegraphics[
height=2.2391in,
width=2.6683in
]%
{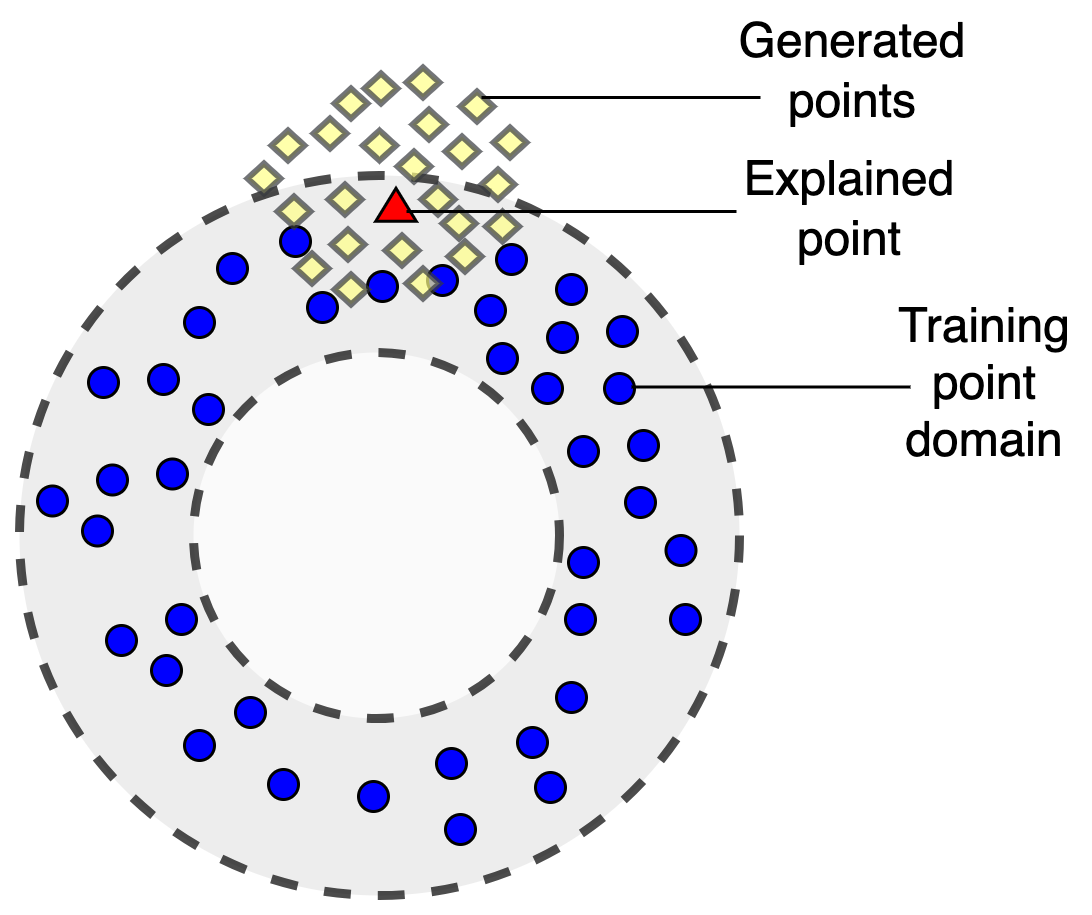}%
\caption{An illustration of a case of out-of-domain data when generated points
may be out of the training point domain}%
\label{f:dual_knn_0}%
\end{center}
\end{figure}

Another important method, which is used for the local as well as global
explanations is SHapley Additive exPlanations (SHAP)
\cite{Lundberg-Lee-2017,Strumbel-Kononenko-2010}. The method is based on
applying game-theoretic Shapley values \cite{Shapley-1953} which can be
interpreted as average marginal contributions of features to the black-box
model prediction. SHAP can be also viewed as a method of the linear
approximation of the black-box model predictions.

One of the important shortcomings of LIME is that it uses the perturbation
technique which may be difficult to implement or may be even incorrect for
some datasets, for example, for images. Moreover, it may provide incorrect
results for high-dimensional data of a complex structure. Moreover, points
generated in accordance with the perturbation technique may be located out of
the training point domain, i.e. these points can be viewed as out-of-domain
(OOD) data. This case is shown in Fig. \ref{f:dual_knn_0} where training
points and generated points are depicted by small circles and by diamonds,
respectively. The explained point is depicted by the triangle. A machine
learning black-box model learned on points from the training domain may
provide quite incorrect predictions for generated points which are outside of
the domain. As a results, the approximating linear function constructed by
using the generated points may be also incorrect.

One of the shortcomings of SHAP is that it is also computationally expensive
when there is a large number of features due to considering all possible
coalitions whose number is $2^{m}$, where $m$ is the number of features.
Therefore, the computational time grows exponentially. Several simplifications
and approximations have been proposed in order to overcome this difficulty
\cite{Strumbel-Kononenko-2010,Strumbelj-Kononenko-11,Strumbel-Kononenko-14,Utkin-Konstantinov-22g}%
. However, they do not cardinally solve the problem of high-dimensional data.
Moreover, there is another difficulty of SHAP, which is rarely mentioned.
According to SHAP, the black-box model prediction is computed for instances
composed from subsets of features and some values of removing features
introduced by using some rules. If to use the example depicted at Fig.
\ref{f:dual_knn_0}, then new instances in SHAP may be located inside or
outside the ring bounding the training data domain where the black-box model
provides incorrect predictions.

In order to partially solve the above problems, we propose a new explanation
method which is based on applying two approaches: the \textit{convex hull} of
training data and the \textit{duality} concept. The convex hull machine
learning methods \cite{Yousefzadeh-21} analyze relationship between a convex
hull of a training set and the decision boundaries for test instances. The
duality is a fundamental concept in various field. We use the dual
representation of data assuming the linear space in the local area around the
explainable instance.

The idea behind the proposed method is very simple. We propose to find the
convex hull of a subset of training data consisting on $K$ instances which are
close to the explainable instance. By using extreme points of the
corresponding convex polytope, each point inside the convex hull can be
expressed through the linear combination of the extreme points. Coefficients
$\mathbf{\lambda}$ of the linear combination are proposed to be regarded as a
new feature vector which determines the corresponding point. They can be
viewed as probabilities defined in the unit simplex of probabilities. Since
the coefficients belong to the unit simplex, then they can be uniformly
generated from the simplex such that each dual feature vector $\mathbf{\lambda
}$ corresponds to the feature vector in the Euclidean space (the feature space
of training data). A generated feature vector in the Euclidean space are
computed through extreme points of the convex hull. As a result, we get a new
dual dataset which generates instances in a local area around the explainable
instance. The surrogate linear model is constructed by using this new dual
dataset whose elements may have a smaller dimension defined by $K$ or by the
number of extreme points of the convex hull. Hence, we get important elements
of the generated vectors of coefficients. Due to the linear representation of
the surrogate (explanation) model, the important features in the Euclidean
space can be simply computed from the important dual coefficients of the
linear combinations by means of solving a simple optimization problem.

Another important idea behind the proposed dual representation is to consider
the example-based explanation. It turns out that the dual explanation
inherently leads to the example-based explanation when we study how each dual
feature $\lambda_{i}$ contributes into predictions. The contribution can be
determined by applying well-known surrogate methods, for example, LIME or the
neural additive model (NAM) \cite{Agarwal-etal-20}, but the corresponding
surrogate models are constructed for features $\mathbf{\lambda}$ but not for
initial features.

For the local explanation, we construct the convex hull by using only a part
of training data. Though the same algorithm can be successfully applied to the
global explanation. In this case, the convex hull covers the entire dataset.

Our contributions can be summarized as follows:

\begin{enumerate}
\item A new feature-based explanation method is proposed. It is based on the
dual representation of datasets such that generation of new instances in
carried out by means of generating points from the uniform distribution in the
unit simplex. In other words, the method replaces the perturbation process of
feature vectors in the Euclidean space by the uniform generation of points in
the unit simplex, which is simpler and is carried out by many well-known
algorithms \cite{Rubinstein-Kroese2008,Smith-Tromble-04}. The generation
resolves the problem of out-of-domain data and reduces the number of
hyperparameters which have to be tuned for perturbing new instances.

\item A new example-based explanation method is proposed. It is again based on
the dual representation of datasets and uses well-known explanation models
NAM, accumulated local effect \cite{Apley-Zhu-20}, the linear regression
model. The explanation method provides shape function which describe
contributions of the dual features into the predictions. In sum, the model
chooses the most influential instances among a certain number of nearest
neighbors for the explained instance.

\item The proposed methods are illustrated by means of numerical experiments
with synthetic and real data. The code of the proposed algorithm can be found
in https://github.com/Kozlov992/Dual-Explanation.
\end{enumerate}

The paper is organized as follows. Related work can be found in Section 2. A
brief introduction to the convex hull, the explanation methods LIME, SHAP, NAM
and example-based methods is given in Section 3. A detailed description of the
proposed approach applied to the feature-based explanation is available in
Section 4. The example-based explanation based on the dual approach is
considered in Section 5. Numerical experiments with synthetic data and real
data studying the feature-based explanation are given in Section 6. Section 7
provides numerical examples illustrating example-based explanation. Concluding
remarks can be found in Section 8.

\section{Related work}

\textbf{Local and global explanation methods.} The requirement of the
black-box model explanation led to development of many explanation methods. A
large part of methods follows from the original LIME method
\cite{Ribeiro-etal-2016}. These methods include ALIME
\cite{Shankaranarayana-Runje-2019}, Anchor LIME \cite{Ribeiro-etal-2018},
LIME-Aleph \cite{Rabold-etal-2019}, SurvLIME \cite{Kovalev-Utkin-Kasimov-20a},
LIME for tabular data \cite{Garreau-Luxburg-2020,Garreau-Luxburg-2020a},
GraphLIME \cite{Huang-Yamada-etal-2020}, etc.

In order to generalize the simple linear explanation surrogate model, several
neural network models, including NAM \cite{Agarwal-etal-20}, GAMI-Net
\cite{Yang-Zhang-Sudjianto-20}, AxNNs \cite{Chen-Vaughan-etal-20} were
proposed. These models are based on applying the GAM
\cite{Hastie-Tibshirani-1990}. Similar explanation models, including
Explainable Boosting Machine \cite{Nori-etal-19}, EGBM
\cite{Konstantinov-Utkin-21}, were developed using the gradient boosting machine.

Another large part of explanation methods is based on the original SHAP method
\cite{Strumbel-Kononenko-2010} which uses Shapley values
\cite{Lundberg-Lee-2017} as measures of the feature contribution into the
black-box model prediction. This part includes FastSHAP \cite{Jethani-etal-21}%
, Kernel SHAP \cite{Lundberg-Lee-2017}, Neighbourhood SHAP
\cite{Ghalebikesabi-etal-21}, SHAFF \cite{Benard-etal-21}, TimeSHAP
\cite{Bento-etal-20}, X-SHAP \cite{Bouneder-etal-20}, ShapNets
\cite{Wang-Wang-Inouye-21}, etc.

Many explanation methods, including LIME and its modifications, are based on
perturbation techniques
\cite{Fong-Vedaldi-2019,Fong-Vedaldi-2017,Petsiuk-etal-2018,Vu-etal-2019},
which stem from the well-known property that contribution of a feature can be
determined by measuring how a prediction changes when the feature is altered
\cite{Du-Liu-Hu-2019}. The main difficulty of using the perturbation technique
is its computational complexity when samples are of the high dimensionality.

Another interesting group of explanation methods, called the example-based
explanation methods \cite{Molnar-2019}, is based on selecting influential
instances from a training set having the largest impact on the predictions and
its comparison with the explainable instance. Several approaches to the
example-based method implementation were considered in
\cite{Adhikari-etal-2019,Cai-Jongejan-etal-19,Chong-Cheung-etal-22,Crabbe-etal-2021,Teso-etal-2021}%
.

In addition to the aforementioned methods, there are a huge number of other
approaches to solving the explanation problem, for example, Integrated
Gradients \cite{Sundararajan-etal-17}, Contrastive Examples
\cite{Dhurandhar-etal-2018}. Detailed surveys of many methods can be found in
\cite{Adadi-Berrada-2018,Arrieta-etal-2020,Bodria-etal-23,Burkart-Huber-21,Carvalho-etal-2019,Islam-etal-22,Guidotti-2019,Li-Xiong-etal-21,Rudin-2019,Rudin-etal-21}%
.

\textbf{Convex hull methods and the convex duality concept}. Most papers
considering the convex hull methods study the relationship between location of
decision boundaries and convex hulls of a training set. The corresponding
methods are presented in
\cite{Chau-Li-Yu-13,Gu-Chung-Wang-00,Nemirko-Dula-21,Nemirko-Dula-21a,Wang-Qiao-etal-13,Yousefzadeh-21}%
. Boundary of the dataset's convex hull are studied in
\cite{Balestriero-etal-21} to discriminate interpolation and extrapolation
occurring for a sample. Efficient algorithms for efficient computation of the
convex hull for training data are presented in \cite{Khosravani-etal-16}.

The concept of duality was also widely used in machine learning models
starting from duality in the support vector machine and its various
modifications \cite{Bennett-Bredensteiner-00,TZhang-02}. This concept was
successfully applied to some types of neural networks
\cite{Ergen-Pilanci-20,Ergen-Pilanci-21}, including GANs \cite{Farnia-Tse-18},
to models dealing with the high-dimensional data \cite{Yao-Zhao-etal-18}.

At the same time, the aforementioned approaches did not applied to explanation
models. Concepts of the convex hull and the convex duality may be a way to
simplify and to improve the explanation models.

\section{Preliminaries}

\subsection{Convex hull}

According to \cite{Rockafellar-70}, a domain produced by a set of instances as
vectors in Euclidean space is convex if a straight line segment that joins
every pair of instances belonging to the set contains a vector belonging to
the domain. A set $\mathcal{S}$ is convex if, for every pair, $\mathbf{u}%
,\mathbf{v}\in \mathcal{S}$, and all $\lambda \in \lbrack0,1]$, the vector
$(1-\lambda)\mathbf{u}+\lambda \mathbf{v}$ belongs to $\mathcal{S}$.

Moreover, if $\mathcal{S}$ is a convex set, then for any $\mathbf{x}%
_{1},\mathbf{x}_{2},...,\mathbf{x}_{t}$ belonging to $\mathcal{S}$ and for any
nonnegative numbers $\lambda_{1},...,\lambda_{t}$ such that $\lambda
_{1}+...+\lambda_{t}=1$, the sum $\lambda_{1}\mathbf{x}_{1}+...+\lambda
_{t}\mathbf{x}_{t}$ is called a convex combination of $\mathbf{x}%
_{1},...,\mathbf{x}_{t}$. The \textit{convex hull} or \textit{convex envelope}
of set $\mathcal{X}$ of instances in the Euclidean space can be defined in
terms of convex sets or convex combinations as the minimal convex set
containing $\mathcal{X}$, or the intersection of all convex sets containing
$\mathcal{X}$, or the set of all convex combinations of instances in
$\mathcal{X}$.

\subsection{LIME, SHAP, NAM and example-based methods}

Let us briefly introduce the most popular explanation methods.

\textit{LIME} \cite{Ribeiro-etal-2016} proposes to approximate a black-box
explainable model, denoted as $f$, with a simple function $g$ in the vicinity
of the point of interest $\mathbf{x}$, whose prediction by means of $f$ has to
be explained, under condition that the approximation function $g$ belongs to a
set of explanation models $G$, for example, linear models. In order to
construct the function $g$, a new dataset consisting of generated points
around $\mathbf{x}$ is constructed with predictions computed be means of the
black-box model. Weights $w_{\mathbf{x}}$ are assigned to new instances in
accordance with their proximity to point $\mathbf{x}$ by using a distance
metric, for example, the Euclidean distance. The explanation function $g$ is
obtained by solving the following optimization problem:
\begin{equation}
\arg \min_{g\in G}L(f,g,w_{\mathbf{x}})+\Phi(g).
\end{equation}

Here $L$ is a loss function, for example, mean squared error, which measures
how the function $g$ is close to function $f$ at point $\mathbf{x}$; $\Phi(g)
$ is the model complexity. A local linear model is the result of the original
LIME such that its coefficients explain the prediction.

Another approach to explaining the black-box model predictions is
\textit{SHAP} \cite{Lundberg-Lee-2017,Strumbel-Kononenko-2010}, which is based
on a concept of the Shapley values \cite{Shapley-1953} estimating
contributions of features to the prediction. If we explain prediction
$f(\mathbf{x}_{0})$ from the model at a local point $\mathbf{x}_{0}$, then the
$i$-th feature contribution is defined by the Shapley value as%
\begin{equation}
\phi_{i}=\sum_{S\subseteq N\backslash \{i\}}\frac{\left \vert S\right \vert
!\left(  \left \vert N\right \vert -\left \vert S\right \vert -1\right)
!}{\left \vert N\right \vert !}\left[  f\left(  S\cup \{i\} \right)  -f\left(
S\right)  \right]  , \label{SHAP_1}%
\end{equation}
where $f\left(  S\right)  $ is the black-box model prediction under condition
that a subset $S$ of the instance $\mathbf{x}_{0}$ features is used as the
corresponding input; $N$ is the set of all features.

It can be seen from (\ref{SHAP_1}) that the Shapley value $\phi_{i}$ can be
regarded as the average contribution of the $i$-th feature across all possible
permutations of the feature set. The prediction $f(\mathbf{x}_{0})$ can be
represented by using Shapley values as follows
\cite{Lundberg-Lee-2017,Strumbel-Kononenko-2010}:
\begin{equation}
f(\mathbf{x}_{0})=\mathbb{E}[f(\mathbf{x})]+\sum_{j=1}^{m}\phi_{j}.
\end{equation}

To generalize LIME, \textit{NAM} was proposed in \cite{Agarwal-etal-20}. It is
based on the generalized additive model of the form $y(\mathbf{x})=g_{1}%
(x_{1})+...+g_{m}(x_{m})$ \cite{Hastie-Tibshirani-1990} and consists of $m$
neural networks such that a single feature is fed to each subnetwork and each
network implements function $g_{i}(x_{i})$, where $g_{i}$ is a univariate
shape function with $\mathbb{E}(g_{i})=0$. All networks are trained jointly
using backpropagation and can learn arbitrarily complex shape functions
\cite{Agarwal-etal-20}.The loss function for training the whole neural network
is of the form:
\begin{equation}
L=\sum_{i=1}^{n}\left(  y_{i}-\sum_{k=1}^{m}g_{k}(x_{k}^{(i)})\right)  ^{2},
\label{SurvNAM_14}%
\end{equation}
where $x_{k}^{(i)}$ is the $k$-th feature of the $i$-th instance; $n$ is the
number of training instances.

The representation of results in NAM in the form of shape functions can be
considered in two ways. On the one hand, the functions are more informative,
and they show how features contribute into a prediction. On the other hand, we
often need to have a single value of the feature contribution which can be
obtained by computing an importance measure from the obtained shape function.

According to \cite{Molnar-2019}, an instance or a set of instances are
selected in \textit{example-based explanation methods} to explain the model
prediction. In contrast to the feature importance explanation (LIME, SHAP),
the example-based methods explain a model by selecting instances from the
dataset and do not consider features or their importance for explaining. In
the context of obtained results, the example-based methods are represented by
influential instances (points from the training set that have the largest
impact on the predictions) and by prototypes (representative instances from
the training data). It should be noted that instances used for explanation may
not belong to a dataset and are combinations of instances from the dataset or
some points in the dataset domain. The well-known method of $K$ nearest
neighbors can be regarded as an example-based explanation method.

\section{Dual explanation}

Let us consider the method for dual explanation. Suppose that there is a
dataset $\mathcal{T}=\{(\mathbf{x}_{1},y_{1}),...,(\mathbf{x}_{t},y_{t})\}$ of
$t$ points $(\mathbf{x}_{i},y_{i})$, where $\mathbf{x}_{i}=(x_{1}%
^{(i)},...,x_{m}^{(i)})\in \mathcal{X}\subset \mathbb{R}^{m}$ is a feature
vector consisting of $m$ features, $y_{i}$ is the observed output for the
feature vector $\mathbf{x}_{i}$ such that $y_{i}\in \mathbb{R}$ in the
regression problem and $y_{i}\in \{1,2,...,C\}$ in the classification problem
with $C$ classes. It is assumed that output $y$ of an explained black-box
model is a function $f(\mathbf{x})$ of an associated input vector $\mathbf{x}$
from $\mathcal{X}$.

In order to explain an instance $\mathbf{x}_{0}\in \mathcal{X}$, an
interpretable surrogate model $g$ for the black-box model $f$ is trained in a
local region around $\mathbf{x}_{0}$. It is carried out by generating a new
dataset $\mathcal{S}$ of $n$ perturbed samples in the vicinity of the point of
interest $\mathbf{x}_{0}$ similarly to LIME. Samples are assigned by weights
$w_{\mathbf{x}}$ in accordance with their proximity to the point $\mathbf{x}$.
By using the black-box model, output values $y$ are obtained as function $f$
of generated instances. As a result, dataset $\mathcal{S}$ consists of $n$
pairs $(\mathbf{x}_{i},f(\mathbf{x}_{i}))$, $i=1,...,n$. Interpretable
surrogate model $g$ is now trained on $\mathcal{S}$. Many explanation methods
like LIME and SHAP are based on applying the linear regression function
\begin{equation}
g(\mathbf{x})=a_{1}x_{1}+...+a_{m}x_{m}=\mathbf{ax}^{\mathrm{T}},
\label{dual_exp_29}%
\end{equation}
as an interpretable model because each coefficient $a_{i}$ in $g$ quantifies
how the $i$-th feature impacts on the prediction. Here $\mathbf{a}%
=(a_{1},...,a_{m})$. It should be noted that the domain of set $\mathcal{S}$
coincides with the domain of set $\mathcal{T}$ in the case of the global explanation.

Let us consider the convex hull $\mathcal{P}$ of a set of $K$ nearest
neighbors of instance $\mathbf{x}_{0}$ in the Euclidean space. The convex hull
$\mathcal{P}$ forms a convex polytope with $d$ vertices or extreme points
$\mathbf{x}_{i}^{\ast}$, $i=1,...,d$. Then each point $\mathbf{x}%
\in \mathcal{P}$ is a convex combination of $d$ extreme points:%
\begin{equation}
\mathbf{x=}\sum_{i=1}^{d}\lambda_{i}\mathbf{x}_{i}^{\ast},\  \text{where
}\lambda_{i}\geq0,\text{ }\sum_{i=1}^{d}\lambda_{i}=1. \label{dual_exp_30}%
\end{equation}

This implies that we can uniformly generate a vector in the unit simplex of
possible vectors $\mathbf{\lambda}$ consisting of $d$ coefficients
$\lambda_{1},...,\lambda_{d}$, denoted $\Delta^{d-1}$. In other words, we can
consider points in the unit simplex $\Delta^{d-1}$ and construct a new dual
dataset $\mathcal{D}=\{(\mathbf{\lambda}^{(1)},z_{1}),...,(\mathbf{\lambda
}^{(n)},z_{n})\}$, which consists of vectors $\mathbf{\lambda}^{(j)}%
=(\lambda_{1}^{(j)},...,\lambda_{d}^{(j)})$, and the corresponding values
$z_{j}$, $j=1,...,n$, computed by using the black-box model $f$ as follows:
\begin{equation}
z_{j}=f\left(  \sum_{i=1}^{d}\lambda_{i}^{(j)}\mathbf{x}_{i}^{\ast}\right)  ,
\label{dual_exp_32}%
\end{equation}
i.e., $z_{j}$ is a prediction of the black-box model when its input is vector
$\sum_{i=1}^{d}\lambda_{i}^{(j)}\mathbf{x}_{i}^{\ast}.$

In sum, we can train the \textquotedblleft dual\textquotedblright \ linear
regression model (the surrogate model) for explanation on dataset
$\mathcal{D}$, which is of the form:%
\begin{equation}
h(\mathbf{\lambda})=b_{1}\lambda_{1}+...+b_{d}\lambda_{d}=\mathbf{b\lambda
}^{\mathrm{T}}, \label{dual_exp_33}%
\end{equation}
where $\mathbf{b}=(b_{1}$,$...,b_{d})$ is the vector of coefficients of the
\textquotedblleft dual\textquotedblright \ linear regression model.

The surrogate model can be trained by means of LIME or SHAP with the dual
dataset $\mathcal{D}$.

Suppose that we have trained the function $h(\mathbf{\lambda})$ and computed
coefficients $b_{1},...,b_{d}$. The next question is how to transform these
coefficients to coefficients $a_{1},...,a_{m}$ which characterize the feature
contribution into the prediction. In the case of the linear regression,
coefficients of function $g(\mathbf{x})=a_{1}x_{1}+...+a_{m}x_{m} $ can be
found from the condition:
\begin{equation}
g\left(  \sum_{i=1}^{d}\lambda_{i}^{(j)}\mathbf{x}_{i}^{\ast}\right)
=h(\mathbf{\lambda}_{j}), \label{dual_exp_42}%
\end{equation}
which has to be satisfied for all generated $\mathbf{\lambda}_{j}$. This
obvious condition means that predictions of the \textquotedblleft
primal\textquotedblright \ surrogate model with coefficients $a_{1},...,a_{m}$
has to coincide with predictions of the \textquotedblleft
dual\textquotedblright \ model.

Introduce a matrix consisting of extreme points%
\begin{equation}
\mathbf{X}=\left(  \mathbf{x}_{i}^{\ast \mathrm{T}}\right)  _{i=1}^{d}.
\label{dual_exp_44}%
\end{equation}

Note that, $\lambda_{i}=1$ and $\lambda_{j}=0$, $j\neq i$, for the $i$-th
extreme point. This implies that the condition (\ref{dual_exp_42}) can be
rewritten as
\begin{equation}
g\left(  \mathbf{x}_{i}^{\ast}\right)  =h(0,...,1_{i},...,0)=b_{i}.
\end{equation}

By using (\ref{dual_exp_29}), we get%
\begin{equation}
g\left(  \mathbf{x}_{i}^{\ast}\right)  =\mathbf{ax}_{i}^{\ast}%
\mathbf{^{\mathrm{T}}=}b_{i}.
\end{equation}

Hence, there holds
\begin{equation}
\mathbf{aX=b}.
\end{equation}

It follows from the above that
\begin{equation}
\mathbf{a=X}^{-1}\mathbf{b},
\end{equation}
where $\mathbf{X}^{-1}$ is the pseudoinverse matrix.

Generally, the vector $\mathbf{a}$ can be computed by solving the following
unconstrained optimization problem:
\begin{equation}
\mathbf{\mathbf{a}}_{opt}=\arg \min_{\mathbf{\mathbf{a}}\in \mathbb{R}^{m}%
}\left \Vert \mathbf{\mathbf{a}X}-\mathbf{b}\right \Vert ^{2}.
\label{dual_exp_46}%
\end{equation}

In the original LIME, perturbed instances are generated around $\mathbf{x}%
_{0}$. One of the important advantages of the proposed dual approach is the
opportunity to avoid generating instances in accordance with a probability
distribution with parameters and to generate only uniformly distributed points
$\mathbf{\lambda}^{(j)}$ in the unit simplex $\Delta^{d-1}$. Indeed, if we
have image data, then it is difficult to perturb pixels or superpixels of
images. Moreover, it is difficult to determine parameters of the generation in
order to cover instances from different classes. According to the dual
representation, after generating vectors $\mathbf{\lambda}^{(j)} $, new
vectors $\mathbf{x}_{j}$ are computed by using (\ref{dual_exp_30}). This is
similar to the mixup method \cite{Zhang-Cisse-etal-18} to some extent that
generates new samples by linear interpolation of multiple samples and their
labels. However, in contrast to the mixup method, the prediction is obtained
as the output of the black-box model (see (\ref{dual_exp_32})), but not as the
convex combination of one-hot label encodings. Another important advantage is
that instances corresponding to the generated set $\mathcal{D}$ are totally
included in the domain of the dataset $\mathcal{T}$. This implies that we do
not get anomalous predictions $f(\mathbf{x}_{i})$ when generated
$\mathbf{x}_{i}$ is far from the domain of the dataset $\mathcal{T}$.

\begin{figure}
[ptb]
\begin{center}
\includegraphics[
height=2.2998in,
width=4.217in
]%
{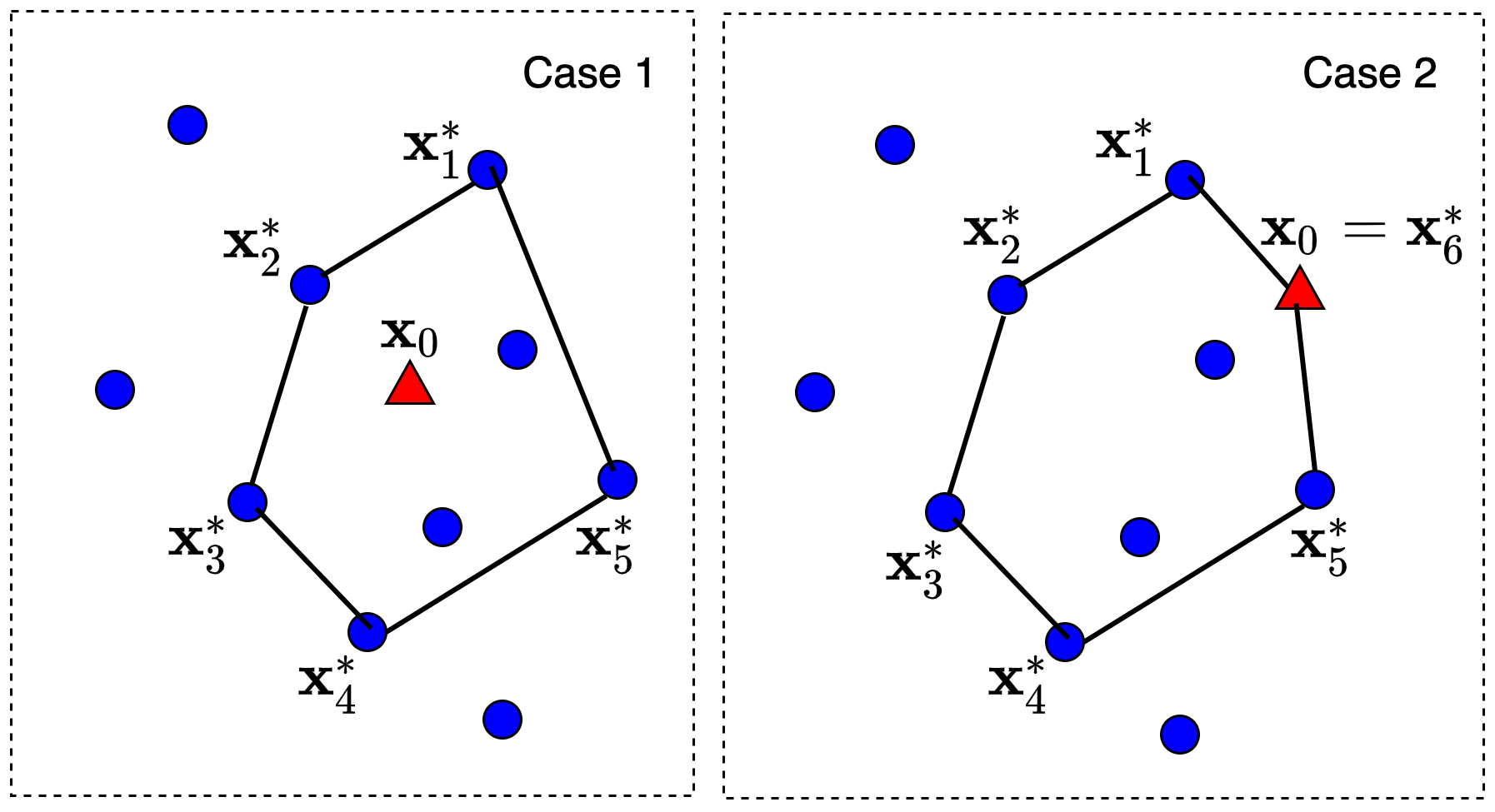}%
\caption{Two cases of the explained point location and the convex polytops
constructed from $K$ nearest neighbors}%
\label{f:dual_knn_1}%
\end{center}
\end{figure}

Another question is how to choose the convex hull of the predefined size and,
hence, how to determine extreme points $\mathbf{x}_{i}^{\ast}$ of the
corresponding convex polytope. The problem is that the convex hull has to
include some number of points from dataset $\mathcal{T}$ and the explained
point $\mathbf{x}_{0}$. Let us consider $K$ nearest neighbors around
$\mathbf{x}_{0}$ from $\mathcal{T}$, where $K$ is a tuning parameter
satisfying condition $K\geq d$. The convex hull is constructed on these $K+1$
points ($K$ points from $\mathcal{T}$ and one point $\mathbf{x}_{0}$). Then
there are $d$ points among $K$ nearest neighbors which define a convex
polytope and can be regarded as its extreme points. It should be noted that
$d$ depends on the dataset analyzed. Fig. \ref{f:dual_knn_1} illustrates two
cases of the explained point location and the convex polytopes constructed
from $K=7$ nearest neighbors. The dataset consists of $10$ points depicted by
circles. A new explained point $\mathbf{x}_{0}$ is depicted by the red
triangle. In Case 1, point $\mathbf{x}_{0}$ lies in the largest convex
polytope with $d=5$ extreme points $\mathbf{x}_{1}^{\ast},...,\mathbf{x}%
_{5}^{\ast}$ constructed from $7$ nearest neighbors. The largest polytope is
taken in order to envelop as large as possible points from the dataset. In
Case 2, point $\mathbf{x}_{0}$ lies outside the convex polytope constructed
from nearest neighbors. Therefore, this point is included into the set of
extreme points and $d\leq K+1$. As a result, we have $d=6$ extreme points
$\mathbf{x}_{1}^{\ast},...,\mathbf{x}_{5}^{\ast},\mathbf{x}_{6}^{\ast
}=\mathbf{x}_{0}$.%

Points $\mathbf{\lambda}^{(j)}$ from the unit simplex $\Delta^{d-1}$ are
randomly selected in accordance with the uniform distribution over the
simplex. This procedure can be carried out by means of generating random
numbers in accordance with the Dirichlet distribution
\cite{Rubinstein-Kroese2008}. There are also different approaches to generate
points from the unit simplex \cite{Smith-Tromble-04}.

Finally, we write Algorithm \ref{alg:Interpr_GBM_0} implementing the proposed method.

\begin{algorithm}
\caption{The dual explanation algorithm} \label{alg:Interpr_GBM_0}

\begin{algorithmic}
[1]\REQUIRE Training set $\mathcal{T}$; the black-box model $f$; explainable
point $\mathbf{x}_{0}$; the number of nearest neighbors $K$

\ENSURE Important features of $\mathbf{x}_{0}$ (vector $\mathbf{a}%
=(a_{1},...,a_{m})^{\mathrm{T}}$ of the linear surrogate model coefficients)

\STATE Determine a set $\mathcal{T}_{K}$ of $K$ nearest neighbors for
$\mathbf{x}_{0}$ adding $\mathbf{x}_{0}$ itself

\STATE Construct the largest convex hull $\mathcal{P}$ of $\mathcal{T}_{K}$

\STATE Find extreme points of $\mathcal{P}$ and their number $d\leq K+1$

\STATE Generate uniformly $n$ points $\mathbf{\lambda}^{(j)}$, $j=1,...,n$,
from the unit simplex $\Delta^{d-1}$

\STATE Find predictions $z_{i}$ of the black-box model in accordance with
associated input $\sum_{i=1}^{d}\lambda_{i}^{(j)}\mathbf{x}_{i}^{\ast}$ for
all $i=1,...,n$

\STATE Construct a new dual dataset $\mathcal{D}=\{(\mathbf{\lambda}%
^{(1)},z_{1}),...,(\mathbf{\lambda}^{(n)},z_{n})\}$

\STATE Train the linear regression (\ref{dual_exp_33}) on dataset
$\mathcal{D}$ and find the vector of coefficients $\mathbf{b}=(b_{1}%
,...,b_{d})^{\mathrm{T}}$

\STATE Find vector $\mathbf{a}$ by solving optimization problem
(\ref{dual_exp_46})
\end{algorithmic}
\end{algorithm}

\begin{figure}
[ptb]
\begin{center}
\includegraphics[
height=3.831in,
width=4.7219in
]%
{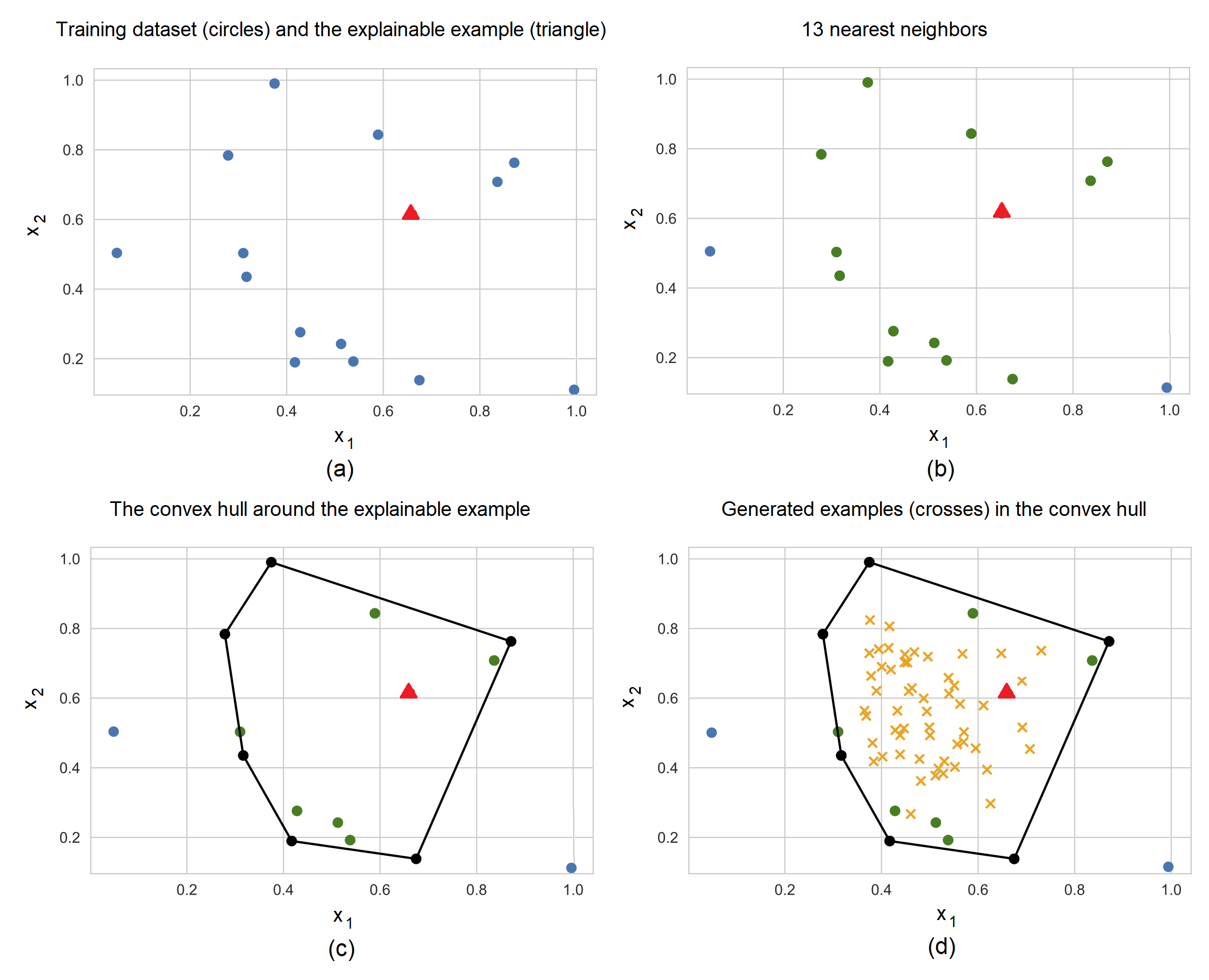}%
\caption{Steps of the algorithm for explanation of a prediction provided by a
black-box model at the point depicted by the small triangle}%
\label{f:steps}%
\end{center}
\end{figure}

Fig. \ref{f:steps} illustrates steps of the algorithm for explanation of a
prediction provided by a black-box model at the point depicted by the small
triangle. Points of the dataset are depicted by small circles. The training
dataset $\mathcal{T}$ and the explained point are shown in Fig. \ref{f:steps}
(a). Fig. \ref{f:steps} (b) shows set $\mathcal{T}_{K}$ of $K=13$ nearest
points such that only two points $(0.05,0.5)$ and $(1.0,0.1)$ from training
set $\mathcal{T}$ do not belong to set $\mathcal{T}_{K}$. The convex hull and
the corresponding extreme points are shown in Fig. \ref{f:steps} (c). Points
uniformly generated in the unit simplex are depicted by means of small crosses
in Fig. \ref{f:steps} (d). It is interesting to point out that the generated
points are uniformly distributed in the unit simplex, but not in the convex
polytope as it is follows from Fig. \ref{f:steps} (d). We uniformly generate
vectors $\mathbf{\lambda}$, but the corresponding vectors $\mathbf{x}$ are not
uniformly distributed in the polytope. One can see from Fig. \ref{f:steps} (d)
that generated points in the initial (primal) feature space tend to be located
in the area where the density of extreme points is largest. This is a very
interesting property of the dual representation. It means that the method
takes into account the concentration of training points and the probability
distribution of the instances in the dataset.%

\begin{figure}
[ptb]
\begin{center}
\includegraphics[
height=1.4529in,
width=4.7307in
]%
{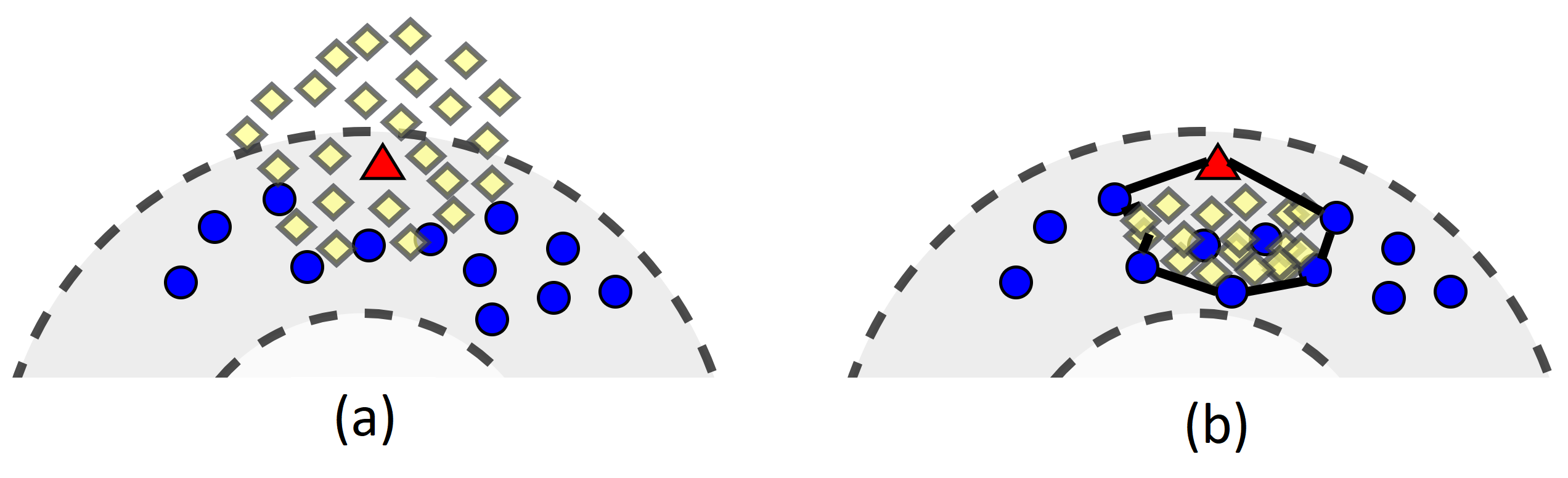}%
\caption{Generated points in the original LIME (a) and in the proposed dual
method (b)}%
\label{f:dual_knn_01}%
\end{center}
\end{figure}

The difference between points generated by means of the original LIME and the
proposed method is illustrated in Fig. \ref{f:dual_knn_01} where the left
picture (Fig. \ref{f:dual_knn_01} (a)) shows a fragment of Fig.
\ref{f:dual_knn_0} and the right picture (Fig. \ref{f:dual_knn_01} (b))
illustrates how the proposed method generates instances.%

\section{Example-based explanation and NAM}

It turns out that the proposed method for the dual explanation inherently
leads to the example-based explanation. An example-based explainer justifies
the prediction on the explainable instance by returning instances related to
it. Let us consider the dual representation (\ref{dual_exp_33}). If we
normalize coefficients $\mathbf{b}=(b_{1},...,b_{d})$ as
\begin{equation}
v_{i}=\frac{b_{i}}{\sum_{j=1}^{d}b_{j}},
\end{equation}
then new coefficients $(v_{1},...,v_{d})$ quantify how extreme points
$(\mathbf{x}_{1}^{\ast},...,\mathbf{x}_{d}^{\ast})$ associated with
$(\lambda_{1},...,\lambda_{d})$ impact on the prediction. The greater the
value of $v_{i}$, the greater contribution of $\mathbf{x}_{i}^{\ast}$ into a
prediction. Hence, the linear combination of extreme points
\begin{equation}
\mathbf{x}=\sum_{i=1}^{d}v_{i}\mathbf{x}_{i}^{\ast} \label{dual_exp_52}%
\end{equation}
allows us to get an instance $\mathbf{x}$ explaining $\mathbf{x}_{0}$.

An outstanding approach considering convex combinations of instances from a
dataset as the example-based explanation was proposed in
\cite{Crabbe-etal-2021}. In fact, we came to the similar example-based
explanation by using the dual representation and constructing linear
regression surrogate model for new variables $(\lambda_{1},...,\lambda_{d}) $.

\begin{figure}
[ptb]
\begin{center}
\includegraphics[
height=3.9277in,
width=3.4273in
]%
{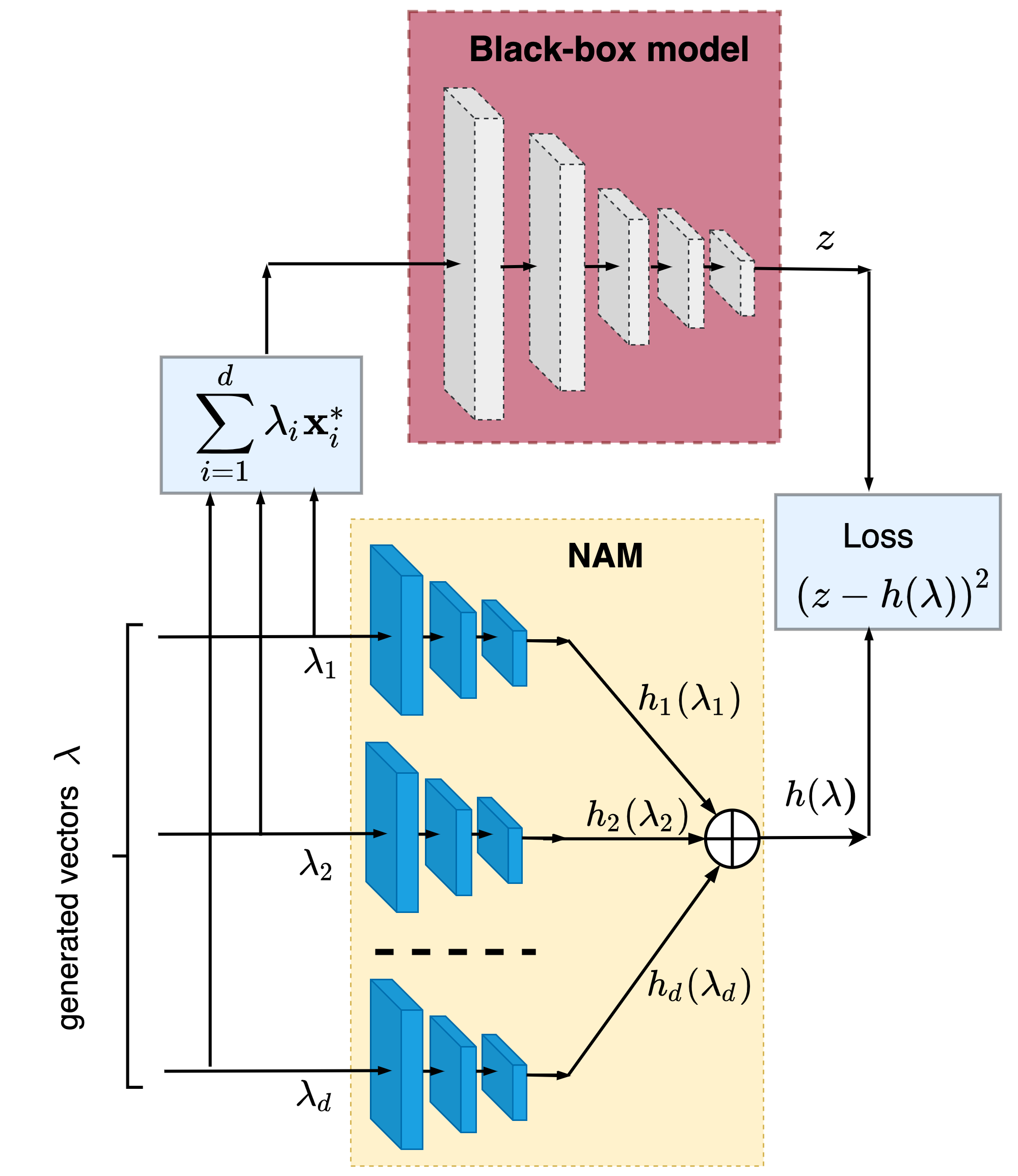}%
\caption{A scheme of training NAM on the generated set of random vectors
$\mathbf{\lambda}$ }%
\label{f:dual_nam}%
\end{center}
\end{figure}

The example-based explanation may be very useful when we apply NAM
\cite{Agarwal-etal-20} for explaining the black-box prediction. By using dual
dataset $\mathcal{D}=\{(\mathbf{\lambda}^{(1)},z_{1}),...,(\mathbf{\lambda
}^{(n)},z_{n})\}$, we train NAM consisting of $d$ subnetworks such that each
subnetwork implements the shape function $h_{i}\left(  \mathbf{\lambda}%
_{i}\right)  $. Fig. \ref{f:dual_nam} illustrates a scheme of training NAM.
Each generated vector $\mathbf{\lambda}$ is fed to NAM such that each its
variable $\lambda_{i}$ is fed to a separate neural subnetwork. For the same
vector $\mathbf{\lambda}$, the corresponding instance $\mathbf{x}$ is computed
by using (\ref{dual_exp_30}), and it is fed to the black-box model. The loss
function for training the whole neural network is defined as the difference
between the output $z$ of the black-box model and the sum of shape functions
$h_{1},...,h_{d}$ implemented by neural subnetworks for the corresponding
vector $\mathbf{\lambda}$, i.e., the loss function $L$ is of the form:%
\begin{equation}
L=\sum_{i=1}^{n}\left(  z_{i}-\sum_{k=1}^{d}h_{k}\left(  \lambda_{k}%
^{(i)}\right)  \right)  ^{2}+\alpha R(\mathbf{w}),
\end{equation}
where $\lambda_{k}^{(i)}$ is the $k$-th element of vector $\mathbf{\lambda
}^{(i)}$; $R$ is a regularization term with the hyperparameter $\alpha$ which
controls the strength of the regularization; $\mathbf{w}$ is the vector of the
neural network training parameters.%

The main difficulty of using the NAM results, i.e. shape functions
$h_{k}\left(  \lambda_{k}\right)  $, is how to interpret the shape functions
for explanation. However, in the context of the example-based explanation,
this difficulty can be simply resolved. First, we study how a shape function
can be represented by a single value characterizing the importance of each
variable $\lambda_{k}$, $k=1,...,d$. The shape function is similar to the
partial dependence plot \cite{Friedman-2001,Molnar-2019} to some extent. The
importance of a variable ($\lambda_{k}$) can be evaluated by studying how
rapidly the shape function, corresponding to the variable, is changed. The
rapid change of the shape function says that small changes of the variable
significantly change the target values ($z$). The above implies that we can
use the importance measure proposed in \cite{Greenwell-etal-18}, which is
defined as the deviation of each unique variable value from the average curve.
In terms of the dual variables, it can be written as:
\begin{equation}
I(\lambda_{k})=\sqrt{\frac{1}{r-1}\sum_{i=1}^{r}\left(  h_{k}\left(
\lambda_{k}^{(i)}\right)  -\frac{1}{r}\sum_{i=1}^{r}h_{k}\left(  \lambda
_{k}^{(i)}\right)  \right)  ^{2}},
\end{equation}
where $r$ is a number of values of each variable $\lambda_{k}$, which are
analyzed to study the corresponding shape function.

Normalized values of the importance measures can be regarded as coefficients
$v_{i}$, $i=1,...,d$, in (\ref{dual_exp_52}), i.e., they show how important
each extreme point or how each extreme point can be regarded as an instance
which explains instance $\mathbf{x}_{0}$.

An additional important advantage of the dual representation is that shape
functions for all variables $\lambda_{k}$, $k=1,...,d$, have the same scale
because all variables are in the interval from $0$ to $1$. This allows us to
compare the importance measures $I(\lambda_{k})$ without the preliminary
scaling which can make results incorrect.

\section{Numerical experiments with the feature-based explanation}

\subsection{Example 1}

First, we consider the following simplest example when the black-box model is
of the form:
\begin{align*}
f(\mathbf{x})  &  =10x_{1}-20x_{2}-2x_{3}+3x_{4}+0x_{5}+0x_{6}+0x_{7}+\xi \\
&  =\mathbf{ax}+\xi,\  \  \xi \sim \mathcal{N}(0,0.1).
\end{align*}

Let us estimate the feature importance by using the proposed dual model. We
generate $n=1000$ points $\mathbf{x}_{i}$, $i=1,...,N$, with components
uniformly distributed in interval $[0,1]$, which are explained. For every
point $\mathbf{x}_{i}$, the dual model with $K=10$ nearest neighbors is
constructed by generating $30$ vectors $\mathbf{\lambda}^{(i)}\in
\mathbb{R}^{7}$ in the unit simplex. By applying Algorithm
\ref{alg:Interpr_GBM_0}, we compute optimal vector $\mathbf{\mathbf{a}}%
^{(i)}=(a_{1},...,a_{7})^{\mathrm{T}}$ for every point $\mathbf{x}_{i}$. We
expect that the mean value $\overline{\mathbf{a}}$ of $\mathbf{\mathbf{a}%
}^{(i)}$ over all $i=1,...,N$, should be as close as possible to the true
vector of coefficients $\mathbf{a} $ forming function $f(\mathbf{x})$. The
corresponding results are shown in Table \ref{t:dual_f_example_1}. It can be
seen from Table \ref{t:dual_f_example_1} that the obtained vector
$\overline{\mathbf{a}}$ is actually close to vector $\mathbf{a}$.%

\begin{table}[tbp] \centering
\caption{Values of the importance measures in Example 1  in accordance with  three explanation approaches:
ALE, LR, NAM}%
\begin{tabular}
[c]{lccccccc}\hline
& $x_{1}$ & $x_{2}$ & $x_{3}$ & $x_{4}$ & $x_{5}$ & $x_{6}$ & $x_{7}$\\ \hline
$\mathbf{a}$ & $10$ & $-20$ & $-2$ & $3$ & $0$ & $0$ & $0$\\ \hline
$\overline{\mathbf{a}}$ & $9.98$ & $-20.01$ & $-2.02$ & $2.97$ & $0.11$ &
$-0.02$ & $0.03$\\ \hline
\end{tabular}
\label{t:dual_f_example_1}%
\end{table}%

\subsection{Example 2}

Let us consider another numerical example where the non-linear black-box model
is investigated. It is of the form:
\[
f(\mathbf{x})=-x_{1}^{2}+2x_{2}+\xi,\  \  \xi \sim \mathcal{N}(0,0.05).
\]

We take $N=400$ and generate two sets of points $\mathbf{x}$. The first set
contains $\mathbf{x}$ whose features are uniformly generated in interval
$[0,1]$. The second set consists of $\mathbf{x}$ whose features are uniformly
generated in interval $[15,16]$. It is interesting to note that feature
$x_{1}$ is more important for the case of the second set because $x_{1}^{2}$
rapidly increases whereas $x_{1}^{2}$ decreases when we consider the first set
and $x_{2}$ is more important in this case.

\begin{figure}
[ptb]
\begin{center}
\includegraphics[
height=1.4832in,
width=2.1629in
]%
{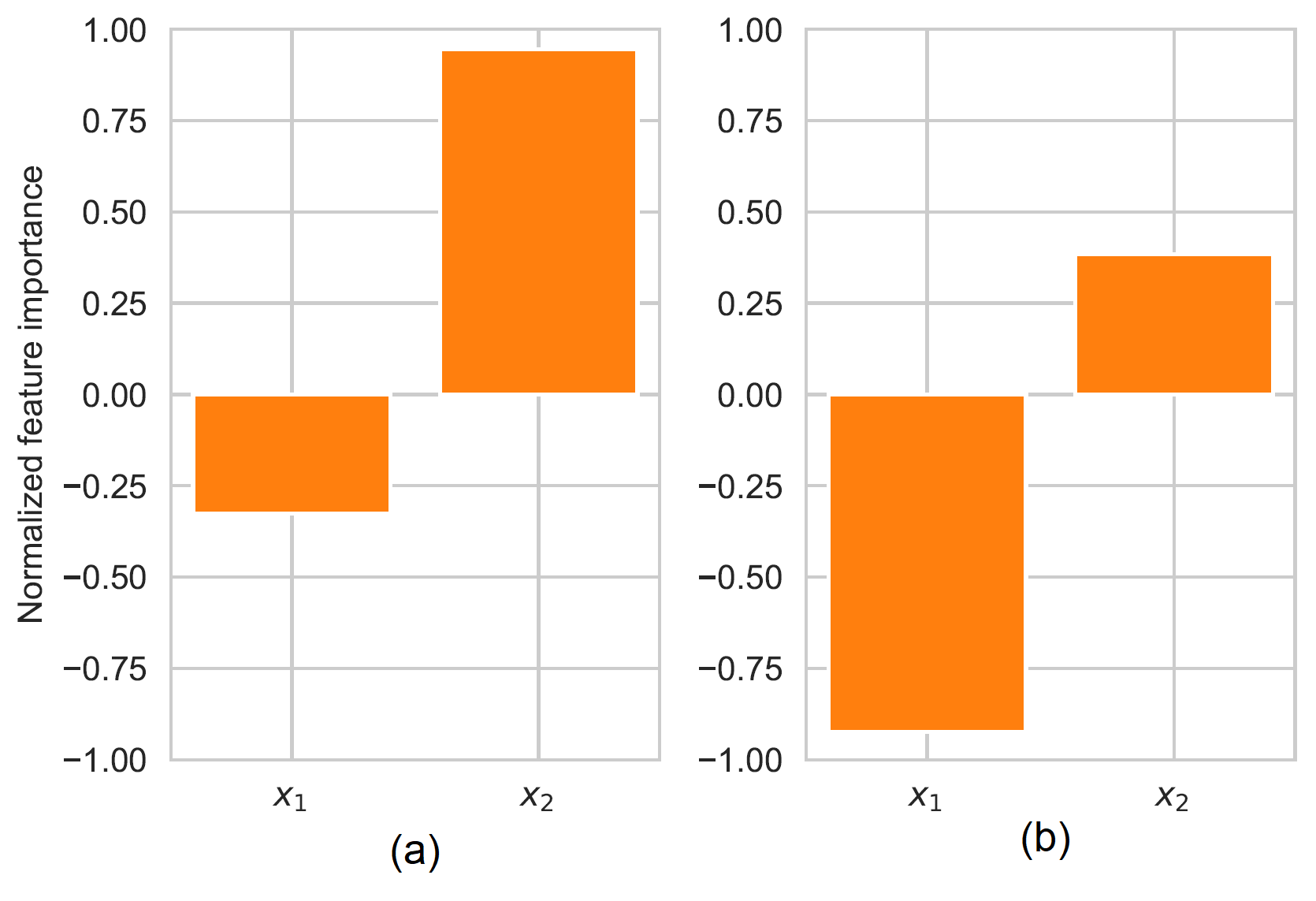}%
\caption{Mean values of the normalized feature importance for two datasets}%
\label{f:examp_f_2}%
\end{center}
\end{figure}

We take $K=6$ and generate $30$ vectors $\mathbf{\lambda}^{(i)}$ uniformly
distributed in the unit simplex for every $\mathbf{x}$ to construct the linear
model $h(\mathbf{\lambda}^{(i)})$. Mean values of the normalized feature
importance obtained for the first and second sets are depicted in Fig.
\ref{f:examp_f_2} (a) and (b), respectively. It can be seen from Fig.
\ref{f:examp_f_2} that the corresponding results completely coincide with the
importance of features considered above for two subsets.%

\subsection{Example 3}

A goal of the following numerical example is to consider a case when we try to
get predictions for points lying outside bounds of data on which the black-box
model was trained as it is depicted in Fig. \ref{f:dual_knn_0}. In this case,
the predictions of generated instances may be inaccurate and can seriously
affect quality of many explanation models, for example, LIME, which uses the
perturbation technique.

The initial dataset consists of $n=400$ feature vectors $\mathbf{x}%
_{1},...,\mathbf{x}_{n}$ such that there holds
\begin{equation}
\mathbf{x}_{i}=\left(
\begin{array}
[c]{c}%
x_{i}^{(1)}\\
x_{i}^{(2)}%
\end{array}
\right)  =\rho \left(
\begin{array}
[c]{c}%
\cos \varphi \\
\sin \varphi
\end{array}
\right)  , \label{dual_exp_110}%
\end{equation}
where parameter $\rho^{2}$ is uniformly distributed in interval $[0,2^{2}]$;
parameter $\varphi$ is uniformly distributed in interval $[0,2\pi]$.

The observed outputs $y_{i}=f(\mathbf{x}_{i})$ are defined as
\begin{equation}
f(\mathbf{x}_{i})=\left(  x_{i}^{(1)}\right)  ^{2}+\left(  x_{i}^{(2)}\right)
^{2}+\xi,\  \  \xi \sim \mathcal{N}(0,0.05). \label{dual_exp_112}%
\end{equation}

We use two black-box models: the KNN regressor with $k=6$ and the random
forest consisting of $100$ decision trees, implemented by means of the Python
Sckit-learn. The above black-box models have default parameters taken from Sckit-learn.

\begin{figure}
[ptb]
\begin{center}
\includegraphics[
height=1.9753in,
width=3.8336in
]%
{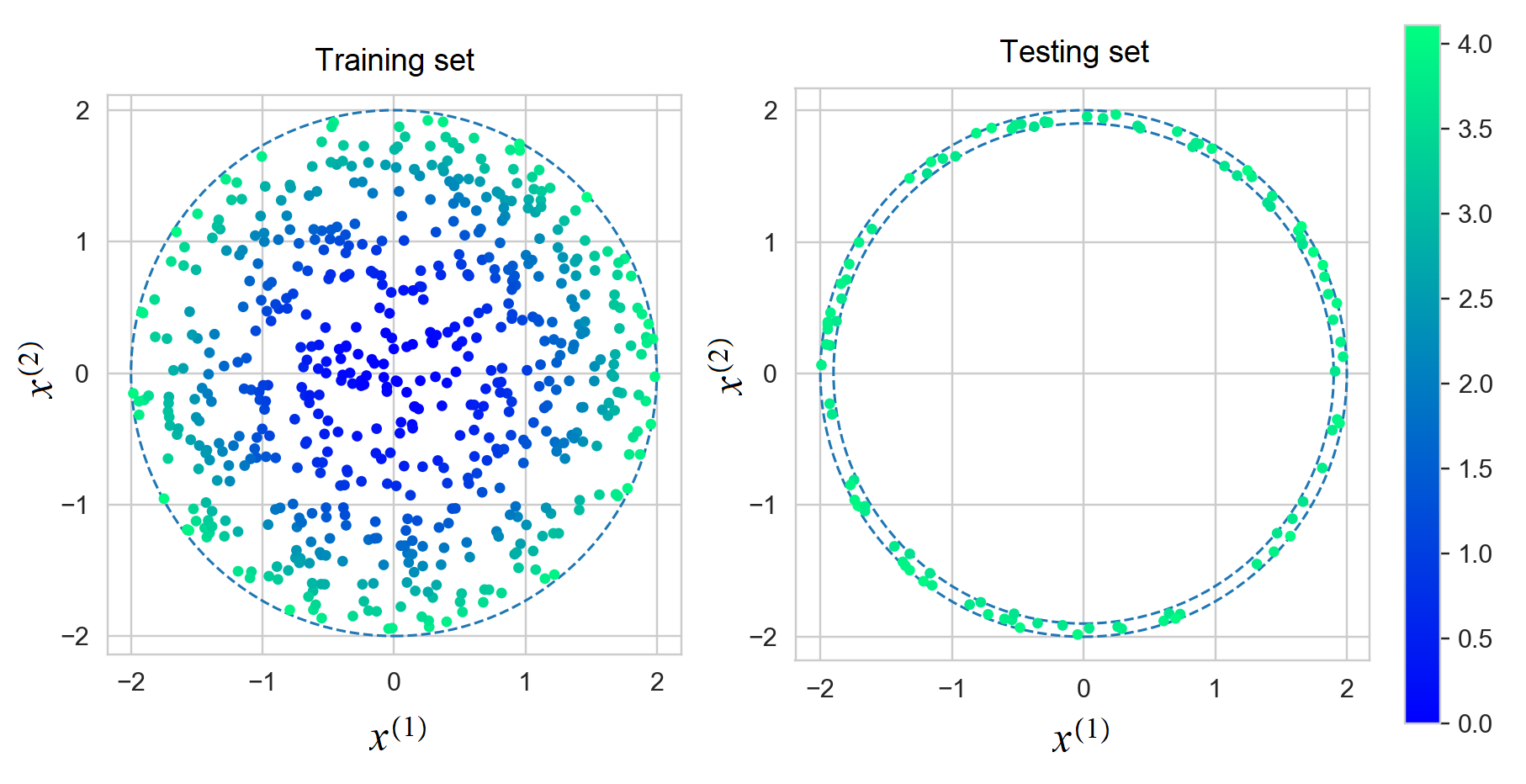}%
\caption{Instances for training the black-box models (the left picture) and
testing points for evaluation of the explanation models (the right picture)}%
\label{f:test_set_bound}%
\end{center}
\end{figure}

We construct the explanation models at $l=100$ testing points $\mathbf{x}%
_{1,test},...,\mathbf{x}_{l,test}$ of the form (\ref{dual_exp_110}), but with
parameters $\rho^{2}$ uniformly distributed in $[1.9^{2},2^{2}]$ and $\varphi$
uniformly distributed in $[0,2\pi]$. It can be seen from the interval of
parameter $\rho$ that a part of generated points can be outside bounds of
training data $\mathbf{x}_{1},...,\mathbf{x}_{n}$. Fig. \ref{f:test_set_bound}
shows the set of instances for training the black-box model and the set of
testing instances for evaluation of the explanation models.%

The dual model is constructed in accordance with Algorithm
\ref{alg:Interpr_GBM_0} using $K=6$ nearest neighbors. We generate $30$ dual
vectors $\mathbf{\lambda}^{(j)}$ to train the dual model. We also use LIME and
generate $30$ points having normal distribution $\mathcal{N}(\mathbf{x}%
_{j,test},\Sigma)$, where $\Sigma=\mathrm{diag}(0.05,0.05)$. Every point has a
weight generated from the normal distribution with parameter $v=0.01$.

In order to compare the dual model and LIME, we use the mean squared error
(MSE) which measures how predictions of the explanation model $g(\mathbf{x})$
are close to predictions of the black-box model $f(\mathbf{x})$ (KNN or the
random forest). It is defined as
\[
MSE=\frac{1}{l}\sum_{j=1}^{l}\left(  f(\mathbf{x}_{j,test})-g(\mathbf{x}%
_{j,test})\right)  ^{2}.
\]

\begin{figure}
[ptb]
\begin{center}
\includegraphics[
height=1.3932in,
width=4.4451in
]%
{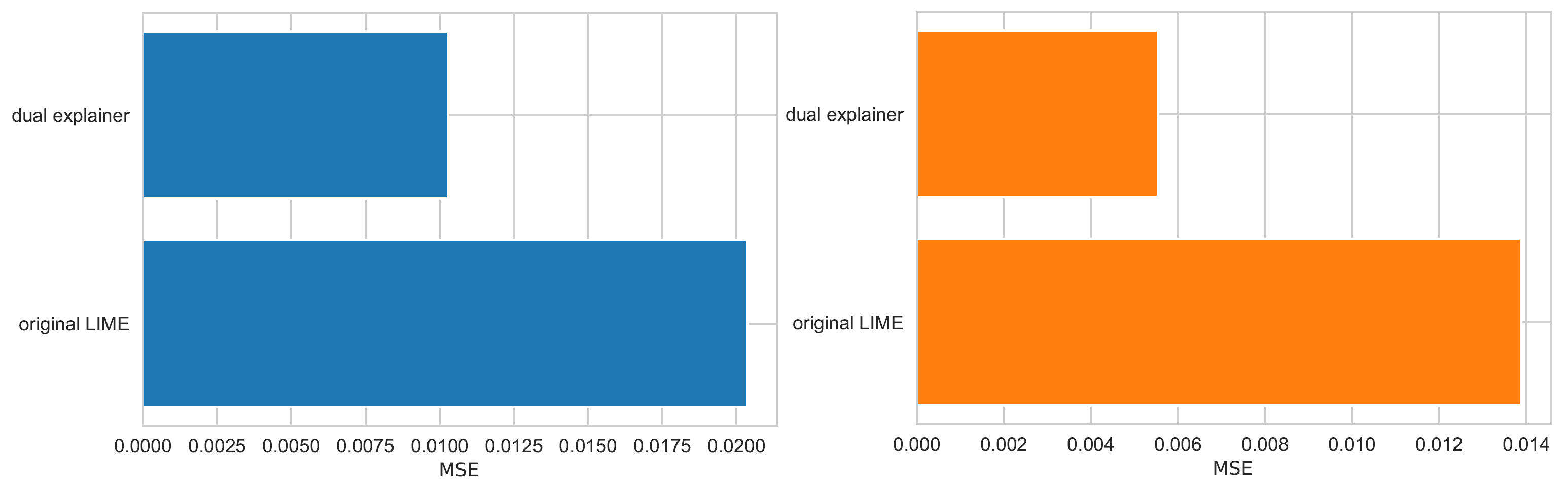}%
\caption{Values of MSE for the dual explanation model and for LIME when KNN
(the left plot) and the random forest (the right plot) are used as black-box
models trained on synthetic data}%
\label{f:example_f_34}%
\end{center}
\end{figure}

Results of numerical experiments with the KNN as a black-box are shown in Fig.
\ref{f:example_f_34} (the left plot). It can be seen from the results that the
dual model provides better results in comparison with LIME because some
generated point in LIME located outside the training domain. Similar results
are shown in Fig. \ref{f:example_f_34} (the right plot) where values of MSE
are depicted when the random forest is used as a black-box model.%

\subsection{Example 4}

Let us perform a similar experiment with real data by taking the dataset
\textquotedblleft Combined Cycle Power Plant Data Set\textquotedblright%
\ (https://archive.ics.uci.edu/ml/datasets/combined+cycle+power+plant)
consisting of 9568 instances having 4 features. We use Z-score normalization
(the mean is $0$ and the standard deviation is $1$) for feature vectors from
the dataset. Two black-box models implemented by using the KNN regressor with
$K=10$ and the random forest regressor consisting of $100$ decision trees. The
testing set consisting of $l=200$ new instances is produced as follows. The
convex hull of the training set in the $4$-dimensional feature space is
determined. Then vertices of the obtained polytope are computed. Two adjacent
vertices $\mathbf{x}_{j_{1}}$ and $\mathbf{x}_{j_{2}}$ are randomly selected.
Value $\lambda$ is generated from the uniform distribution on the unit
interval. A new testing instance $\mathbf{x}_{j,test}$ is obtained as
$\mathbf{x}_{j,test}=\lambda \mathbf{x}_{j_{1}}+(1-\lambda)\mathbf{x}_{j_{2}}$.
Then we again select adjacent vertices and repeat the procedure for computing
testing instances $l$ times. As a result, we get the testing set
$\mathbf{x}_{j,test}$, $j=1,...,l$.

The dual model is constructed in accordance with Algorithm
\ref{alg:Interpr_GBM_0} using $K=10$ nearest neighbors. We again generate $30$
dual vectors $\mathbf{\lambda}^{(j)}$ to train the dual model. We also use
LIME and generate $30$ points having normal distribution $\mathcal{N}%
(\mathbf{x}_{j,test},\Sigma)$, where $\Sigma=\mathrm{diag}%
(0.05,0.05,0.05,0.05)$. Every point has a weight generated from the normal
distribution with parameter $v=0.5$.

\begin{figure}
[ptb]
\begin{center}
\includegraphics[
height=1.4373in,
width=4.4927in
]%
{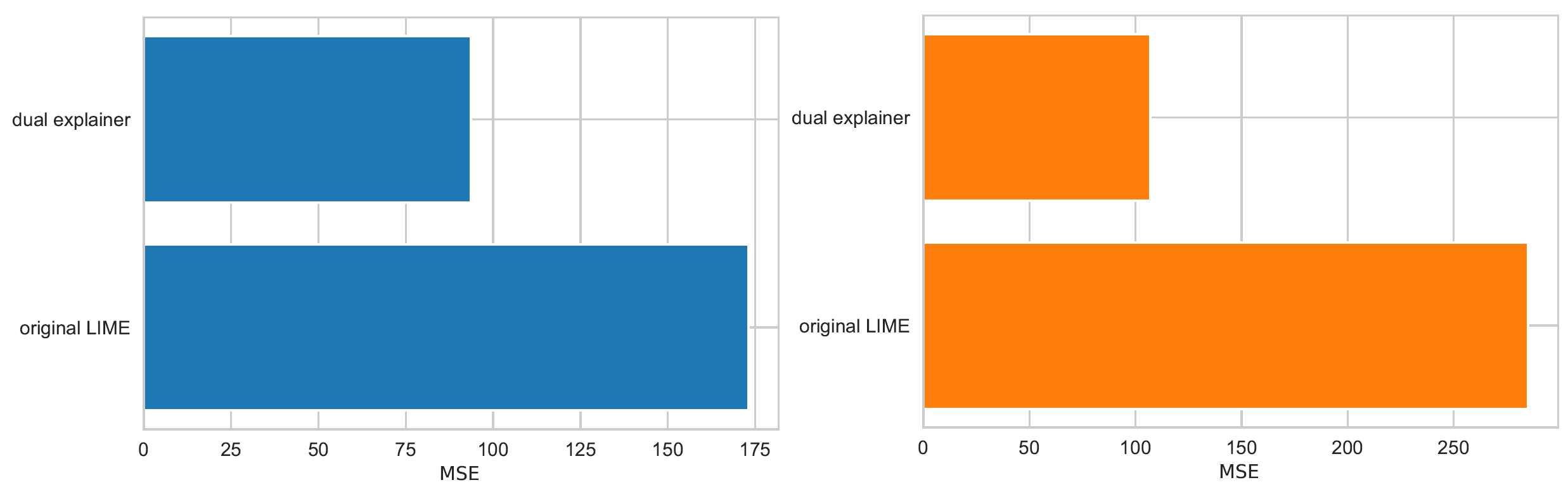}%
\caption{Values of MSE for the dual explanation model and for LIME when KNN
(the left plot) and the random forest (the right plot) are used as black-box
models trained on dataset \textquotedblleft Combined Cycle Power Plant Data
Set\textquotedblright}%
\label{f:examp_f_67}%
\end{center}
\end{figure}

The left plot in Fig. \ref{f:examp_f_67} shows the MSE measures obtained for
the dual model and LIME when the KNN regressor as a black-box model is used.
The same MSE measures, obtained when the random forest as a black-box model is
used, are shown in the right plot in Fig. \ref{f:examp_f_67}. One can see from
Fig. \ref{f:examp_f_67} that the dual models outperform LIME.%

\section{Numerical experiments with the example-based explanation}

\subsection{Example 1}

We start from the synthetic instances illustrating the dual example-based
explanation when NAM is used. Suppose that the explained instance
$\mathbf{x}_{0}$ belongs to a polytope with six vertices $\mathbf{x}%
_{1},...,\mathbf{x}_{6}$ ($d=6$). The black-box model is a function
$f(\mathbf{x})$ such that
\begin{align}
f(\mathbf{x}) &  =f\left(  \sum_{k=1}^{6}\lambda_{k}\mathbf{x}_{k}\right)
=h(\mathbf{\lambda})=h(\lambda_{1},...,\lambda_{6})\nonumber \\
&  =15\lambda_{1}+22\lambda_{2}+0\lambda_{3}+40(1-\lambda_{4})\sin
(3.14\cdot \lambda_{4})+0\lambda_{5}+0\lambda_{6}.\label{dual_exp_120}%
\end{align}

$n=2000$ vectors $\mathbf{\lambda}^{(i)}\in \mathbb{R}^{6}$, $i=1,...,n$, are
uniformly generated in the unit simplex $\Delta^{6-1}$. For each point
$\mathbf{\lambda}^{(i)}$, the corresponding prediction $z_{i}$ is computed by
using the black-box function $h(\mathbf{\lambda})$. NAM is trained with the
learning rate $0.0005$, with hyperparameter $\alpha=10^{-4}$, the number of
epochs is $300$ and the batch size is $128$.

In order to determine the normalized values of the importance measures
$I(\lambda_{i})$, $i=1,...,6$, we use three approaches. The first one is to
apply the method called accumulated local effect (ALE) \cite{Apley-Zhu-20},
which describes how features influence the prediction of the black-box model
on average. The second approach is to construct the linear regression model
(LR) by using the generated points and their predictions obtained by means of
the black-box model. The third approach is to use NAM.

The corresponding normalized values of the importance measures for
$\lambda_{1},...,\lambda_{6}$ obtained by means of ALE, LR and NAM are shown
in Table \ref{t:dual_example_1}. It should be noted that the importance
measure $I(\lambda_{i})$ can be obtained only for NAM and ALE. However,
normalized coefficients of LR can be interpreted in the same way. Therefore,
we consider results of these models jointly in all tables. One can see from
Table \ref{t:dual_example_1} that all methods provide similar relationships
between the importance measures $I(\lambda_{1})$, $i=1,...,6$. However, LR
provides rather large values of $I(\lambda_{3})$, $I(\lambda_{5})$,
$I(\lambda_{6})$, which do not correspond to the zero-valued coefficients in
(\ref{dual_exp_120}).

\begin{figure}
[ptb]
\begin{center}
\includegraphics[
height=2.8257in,
width=4.772in
]%
{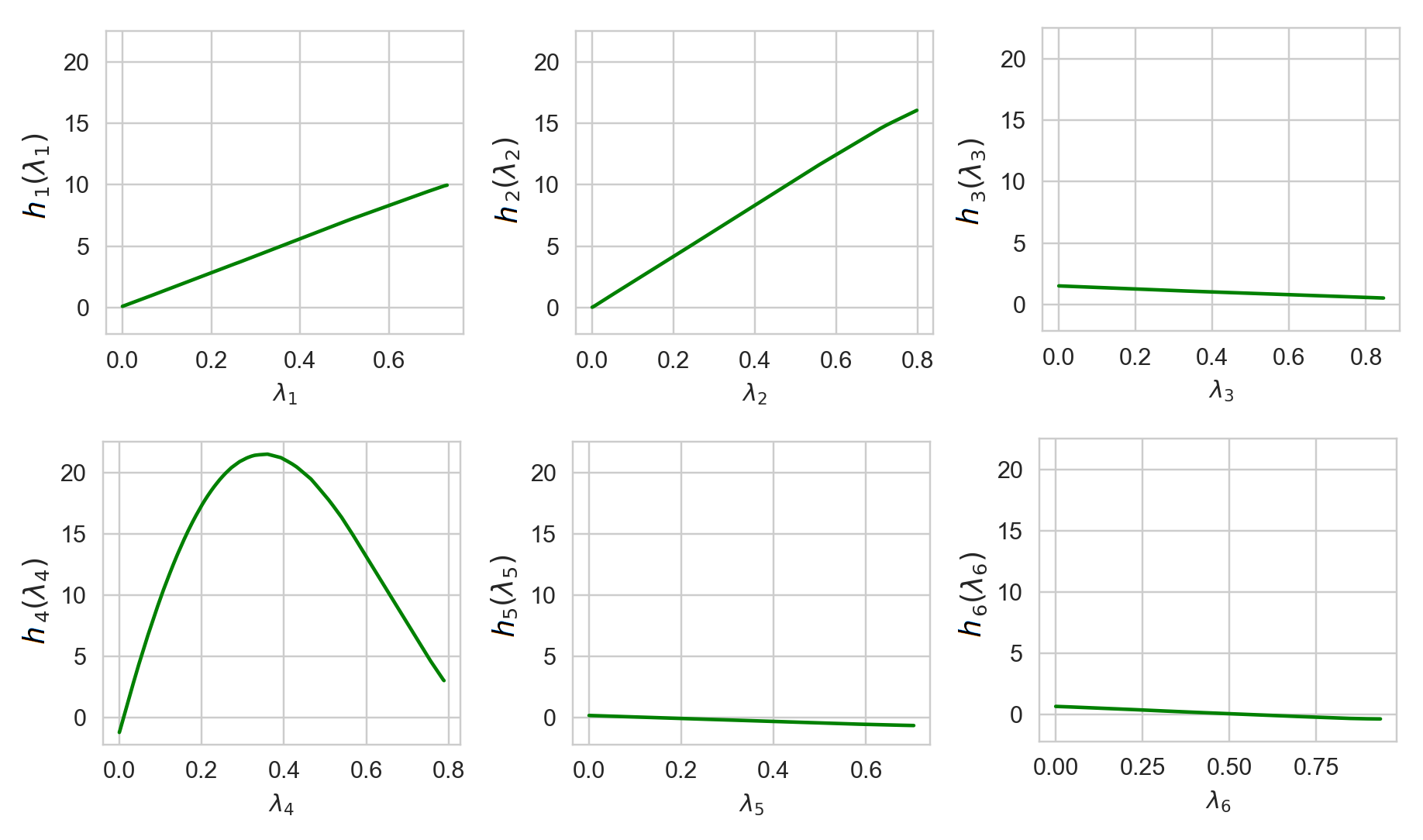}%
\caption{Six shape functions obtained in Example 1 for the example-based
explanation }%
\label{f:g_all_1}%
\end{center}
\end{figure}

Shape functions illustrating how functions of the generalized additive model
depend on $\lambda_{i}$ are shown in Fig. \ref{f:g_all_1}. It can be clearly
seen from Fig. \ref{f:g_all_1} that the largest importance $\lambda_{2}$ and
$\lambda_{4}$ have the highest importance. This implies that the explained
instance is interpreted by the fourth and the second nearest instances.%

\begin{table}[tbp] \centering
\caption{Values of the importance measures in Example 1  in accordance with  three explanation approaches:
ALE, LR, NAM}%
\begin{tabular}
[c]{lcccccc}\hline
& \multicolumn{6}{c}{Importance measures}\\ \hline
& $I(\lambda_{1})$ & $I(\lambda_{2})$ & $I(\lambda_{3})$ & $I(\lambda_{4})$ &
$I(\lambda_{5})$ & $I(\lambda_{6})$\\ \hline
ALE & $0.172$ & $0.259$ & $0.000$ & $0.569$ & $0.000$ & $0.000$\\ \hline
LR & $0.182$ & $0.245$ & $0.054$ & $0.405$ & $0.062$ & $0.052$\\ \hline
NAM & $0.157$ & $0.238$ & $0.012$ & $0.569$ & $0.012$ & $0.012$\\ \hline
\end{tabular}
\label{t:dual_example_1}%
\end{table}%
%

\subsection{Example 2}

Suppose that the explainable instance $\mathbf{x}_{0}$ belongs to a polytope
with four vertices $\mathbf{x}_{1}^{\ast},...,\mathbf{x}_{4}^{\ast}$ ($d=4$).
The black-box model is a function $f(\mathbf{x})$ such that
\[
h(\mathbf{\lambda})=\lambda_{1}^{2}+\lambda_{1}\lambda_{2}-\lambda_{3}%
\lambda_{4}+\lambda_{4}.
\]

$n=1000$ points $\mathbf{\lambda}^{(i)}\in \mathbb{R}^{4}$, $i=1,...,n$, are
uniformly generated in the unit simplex $\Delta^{4-1}$. For each point
$\mathbf{\lambda}^{(i)}$, the corresponding prediction $z_{i}$ is computed by
using the black-box function $h(\mathbf{\lambda})$. NAM is trained with the
learning rate $0.0005$, with hyperparameter $\alpha=10^{-6}$, the number of
epochs is $300$ and the batch size is $128$.

\begin{figure}
[ptb]
\begin{center}
\includegraphics[
height=2.6639in,
width=2.8829in
]%
{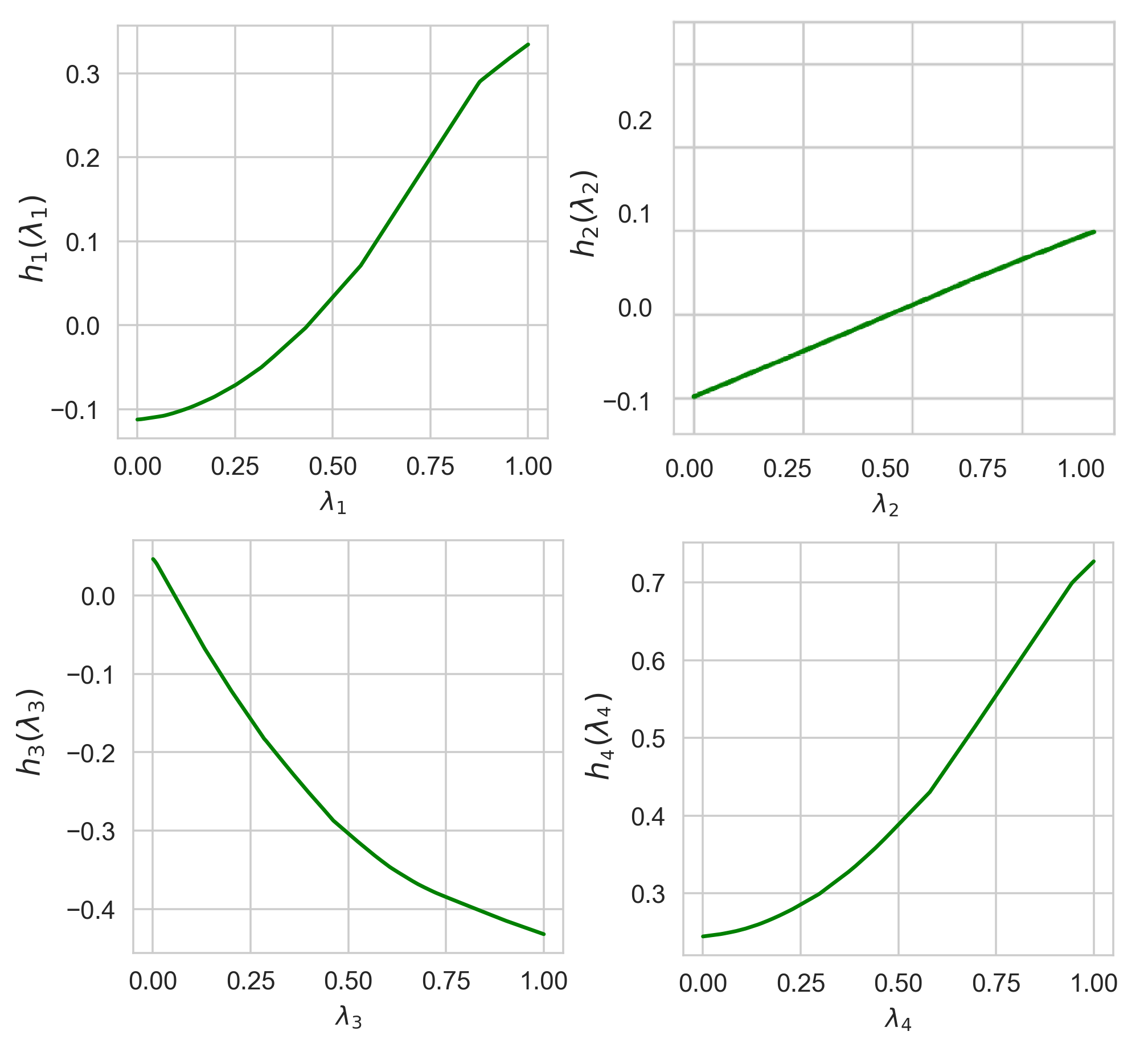}%
\caption{Four shape functions obtained in Example 2 for the example-based explanation }
\label{f:g_all_2}%
\end{center}
\end{figure}

Normalized values of $I(\lambda_{i})$ obtained by means of ALE, LR and NAM are
shown in Table \ref{t:dual_example_2}. It can be seen from Table
\ref{t:dual_example_2} that the obtained importance measures correspond to the
intuitive consideration of the expression for $h(\mathbf{\lambda})$. The
corresponding shape functions for all features are shown in Fig.
\ref{f:g_all_2}.%

\begin{table}[tbp] \centering
\caption{Values of the importance measures in Example 2  in accordance with  three explanation approaches:
ALE, LR, NAM}%
\begin{tabular}
[c]{lcccc}\hline
& \multicolumn{4}{c}{Importance measures}\\ \hline
& $I(\lambda_{1})$ & $I(\lambda_{2})$ & $I(\lambda_{3})$ & $I(\lambda_{4}%
)$\\ \hline
ALE & $0.392$ & $0.087$ & $0.089$ & $0.432$\\ \hline
LR & $0.357$ & $0.081$ & $0.112$ & $0.450$\\ \hline
NAM & $0.306$ & $0.134$ & $0.202$ & $0.358$\\ \hline
\end{tabular}
\label{t:dual_example_2}%
\end{table}%
%

\subsection{Example 3}

Suppose that the explained instance $\mathbf{x}_{0}$ belongs to a polytope
with three vertices $\mathbf{x}_{1}^{\ast},\mathbf{x}_{2}^{\ast}%
,\mathbf{x}_{3}^{\ast}$ ($d=3$):
\[
\mathbf{x}_{1}^{\ast}=(-1,-1)^{\mathrm{T}},\  \mathbf{x}_{2}^{\ast
}=(0,2)^{\mathrm{T}},\  \mathbf{x}_{3}^{\ast}=(1,0)^{\mathrm{T}}.\
\]

The black-box model has the following function of two features $x^{(1)}$ and
$x^{(2)}$:
\[
f(\mathbf{x})=0.7\cdot \mathrm{sign}(x^{(1)})+\mathrm{sign}(x^{(2)})
\]

\begin{figure}
[ptb]
\begin{center}
\includegraphics[
height=2.218in,
width=3.3191in
]%
{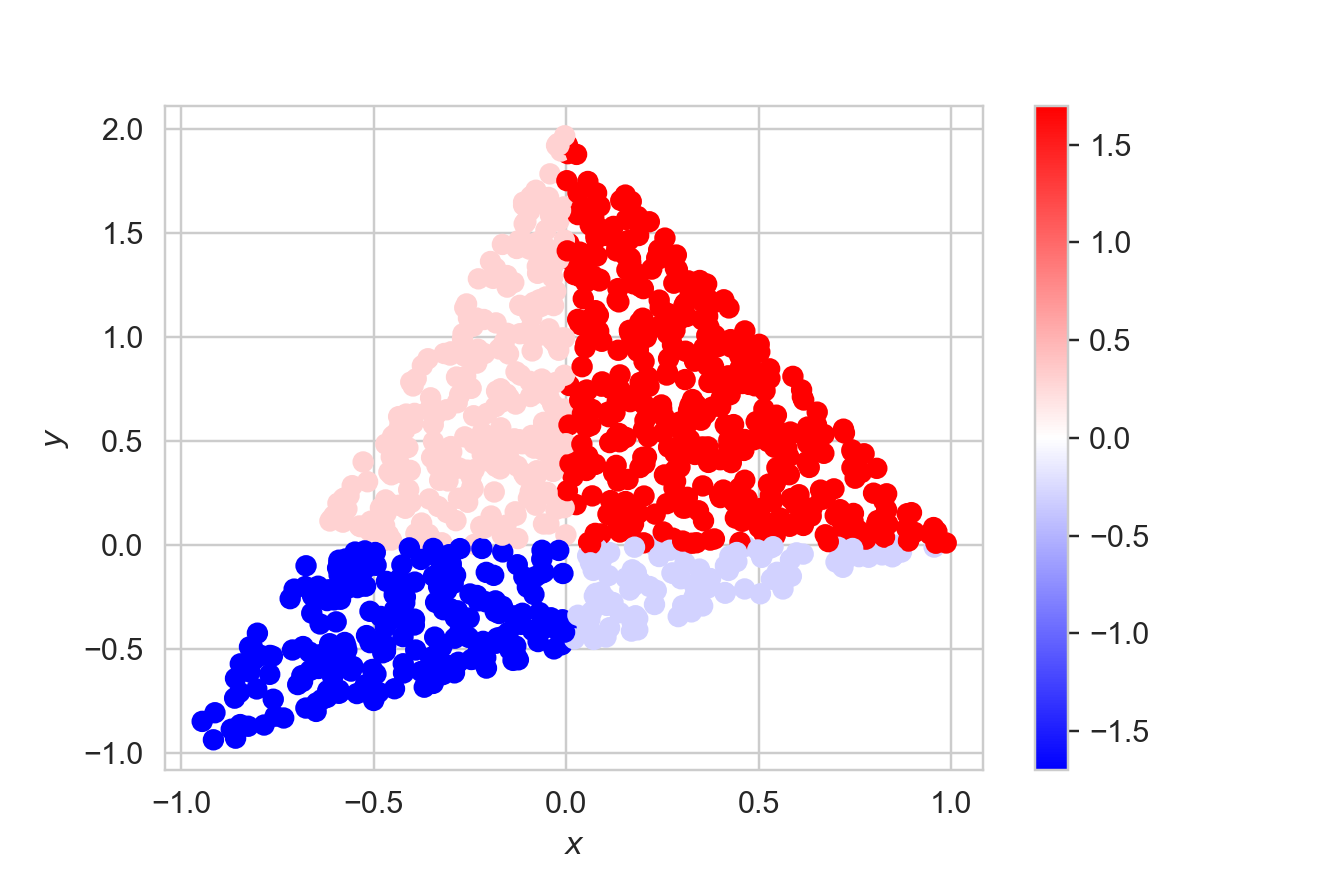}%
\caption{The dataset of vectors $\mathbf{x}$ and the corresponding values of
$f(\mathbf{x})$ for Example 3}%
\label{f:data_set_3}%
\end{center}
\end{figure}

We generate $n=1000$ points $\mathbf{\lambda}^{(i)}\in \mathbb{R}^{3}$,
$i=1,...,n$, which are uniformly generated in the unit simplex $\Delta^{3-1}$.
These points correspond to $n$ vectors $\mathbf{x}_{i}\in \mathbb{R}^{2}$
defined as
\[
\mathbf{x}_{i}=\mathbf{\lambda}_{1}^{(i)}\cdot(-1,-1)^{\mathrm{T}%
}+\mathbf{\lambda}_{2}^{(i)}\cdot(0,2)^{\mathrm{T}}+\mathbf{\lambda}_{3}%
^{(i)}\cdot(1,0)^{\mathrm{T}}%
\]
with the corresponding values of $f(\mathbf{x}_{i})$ and shown in Fig.
\ref{f:data_set_3}. It can be seen from Fig. \ref{f:data_set_3} that this
example can be regarded as a classification task with $4$ classes. Parameters
of experiments are the same as in the previous examples, but $\alpha=0$.

\begin{figure}
[ptb]
\begin{center}
\includegraphics[
height=1.8047in,
width=5.2267in
]%
{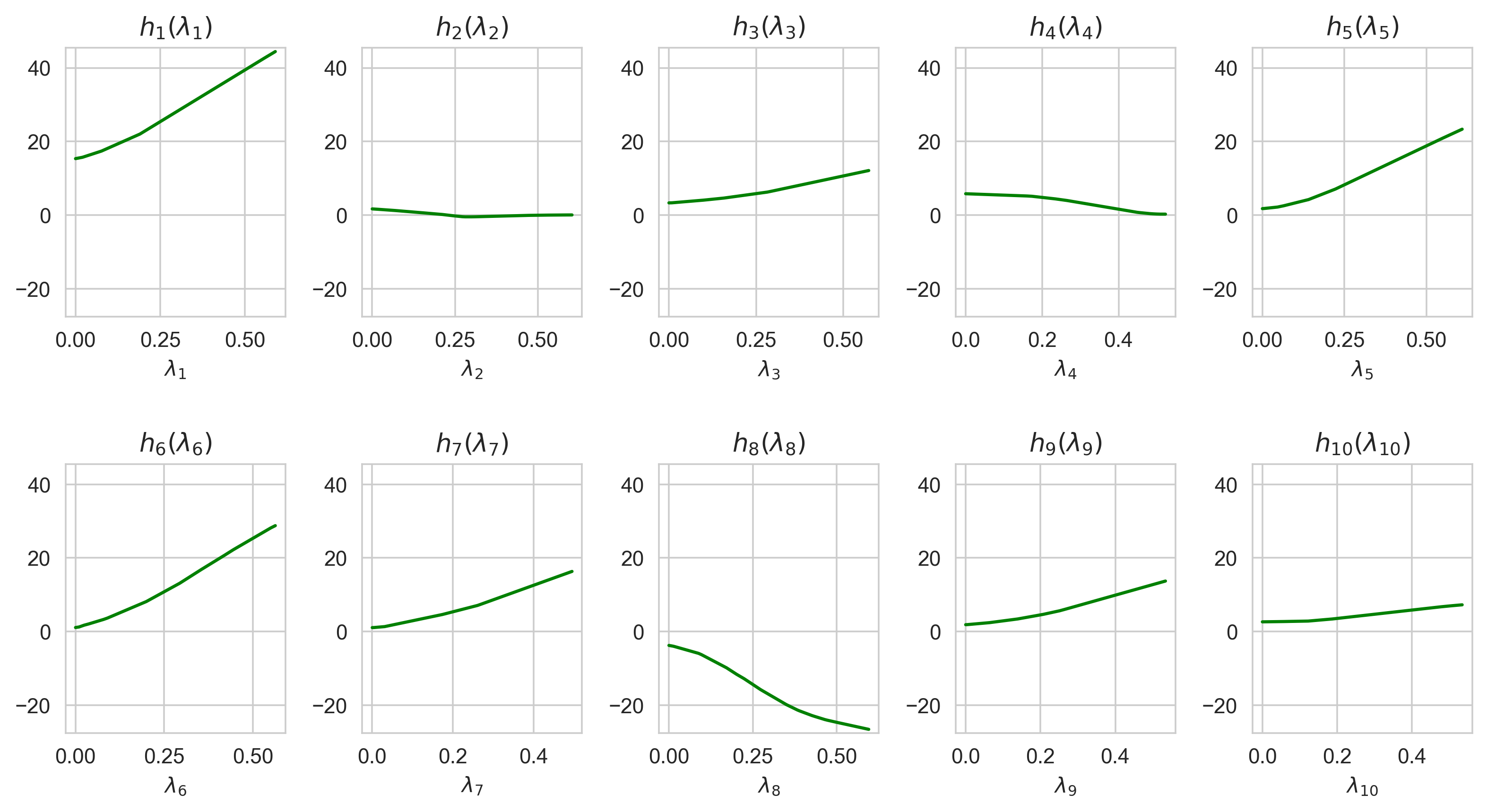}%
\caption{Three shape functions obtained in Example 3 for the example-based
explanation }%
\label{f:g_all_3}%
\end{center}
\end{figure}

Normalized values of $I(\lambda_{i})$ obtained by means of ALE, LR and NAM are
shown in Table \ref{t:dual_example_3}. It can be seen from Table
\ref{t:dual_example_3} that the obtained importance measures correspond to the
intuitive consideration of the expression for $h(\mathbf{\lambda})$. The
corresponding shape functions for all features are shown in Fig.
\ref{f:g_all_3}.%

%

\begin{table}[tbp] \centering
\caption{Values of the importance measures in Example 3 in accordance with three explanation approaches:
ALE, LR, NAM}%
\begin{tabular}
[c]{lccc}\hline
& \multicolumn{3}{c}{Importance measures}\\ \hline
& $I(\lambda_{1})$ & $I(\lambda_{2})$ & $I(\lambda_{3})$\\ \hline
ALE & $0.411$ & $0.395$ & $0.194$\\ \hline
LR & $0.430$ & $0.310$ & $0.260$\\ \hline
NAM & $0.499$ & $0.338$ & $0.163$\\ \hline
\end{tabular}
\label{t:dual_example_3}%
\end{table}%
%

\section{Conclusion}

Let us analyze advantages and limitations of the proposed methods. First, we
consider advantages.

\begin{enumerate}
\item One of the important advantaged is that the proposed methods allow us to
replace the perturbation process of feature vectors in the Euclidean space by
the uniform generation of points in the unit simplex. Indeed, the perturbation
of feature vectors requires to define several parameters, including
probability distributions of generation for every feature, parameters of the
distributions, etc. The cases depicted in Fig. \ref{f:dual_knn_0} may lead to
incorrect predictions and to an incorrect surrogate model. Moreover, if
instances are images, then it is difficult to correctly perturb them. Due to
the proposed method, the perturbation of feature vectors is avoided, and it is
replaced with uniform generation in the unit simplex, which is simple. The
dual approach can be applied to the feature-based explanation as well as to
the example-based explanation.

\item The dual representation of data can have a smaller dimension than the
initial instances. It depends on $K$\ nearest neighbors around the explained
instance. As a result, the constructed surrogate dual model can be simpler
than the model trained on the initial training set.

\item The dual approach can be also adapted to SHAP to generate the removed
features in a specific way.

\item The proposed methods are flexible. We can change the size of the convex
hull by changing the number $K$. It can be applied to different explanation
models, for example, to LIME, SHAP, NAM. The method can be applied to the
local as well as global explanations.
\end{enumerate}

In spite of many advantages of the dual approach, we have to note also its limitations:

\begin{enumerate}
\item The advantage of the smaller dimensionality in the dual representation
is questionable for the feature-based explanation. If we take a number of
extreme points smaller than the data dimensionality, then we restrict the set
of generated primal points by some subspace of the initial feature space. This
can be a reason of incorrect results. Ways to overcome this difficulty is an
interesting direction for further research. However, this limitation does not
impact on the example-based explanation because we actually extend the mixup
method and try to find influential instances among nearest neighbors.

\item Another problem is that calculation of vertices of the largest convex
hull is a computationally hard problem. This problem does not take place for
the example-based explanation when the number of nearest neighbors smaller
than the initial data dimensionality.
\end{enumerate}

In spite of the above limitations, the proposed approach has many interesting
properties and can be regarded as the first step for developing various
algorithms using dual representation.

\bibliographystyle{unsrt}
\bibliography{Boosting,Convex,Expl_Attention,Explain,Explain_med,Lasso,MYBIB,MYUSE,Robots}

\end{document}